\pdfoutput=1

\documentclass[11pt]{article}

\usepackage[preprint]{neurips_2023}
\usepackage{xcolor}
\usepackage[colorlinks, linkcolor={blue!60!black}, citecolor={blue!60!black}, urlcolor={blue!60!black}]{hyperref}

\usepackage{times}
\usepackage{latexsym}

\usepackage{colortbl}

\usepackage{multirow}

\usepackage[T1]{fontenc}
\usepackage{booktabs}
\usepackage{subcaption}

\usepackage[utf8]{inputenc}

\usepackage{microtype}

\usepackage{inconsolata}
\usepackage{soul}
\newcommand{\hlred}[1]{\sethlcolor{red!35}\hl{#1}}

\usepackage{graphicx}

\usepackage[font=small,labelfont=bf]{caption}

\usepackage{enumitem}
\setlist[description]{font=\normalfont\itshape}

\title{Hallucination-Free? Assessing the Reliability of Leading AI Legal Research Tools}

\author{%
\begin{tabular}{ccc}
\begin{tabular}[t]{c}
Varun Magesh\thanks{Equal contribution.} \\
{\normalfont Stanford University} \\
\end{tabular} &
\begin{tabular}[t]{c}
Faiz Surani\footnotemark[1] \\
{\normalfont Stanford University} \\
\end{tabular} &
\begin{tabular}[t]{c}
Matthew Dahl \\
{\normalfont Yale University} \\
\end{tabular} \\ \addlinespace[4ex]
\begin{tabular}[t]{c}
Mirac Suzgun \\
{\normalfont Stanford University} \\
\end{tabular} &
\begin{tabular}[t]{c}
Christopher D. Manning \\
{\normalfont Stanford University} \\
\end{tabular} &
\begin{tabular}[t]{c}
Daniel E. Ho\thanks{Corresponding author: \url{deho@stanford.edu}.} \\
{\normalfont Stanford University} \\
\end{tabular}
\end{tabular}
}

\begin{document}
\maketitle
\setcounter{footnote}{0}

\begin{abstract}
Legal practice has witnessed a sharp rise in products incorporating artificial intelligence (AI).
Such tools are designed to assist with a wide range of core legal tasks, from search and summarization of caselaw to document drafting. But the large language models used in these tools are prone to ``hallucinate,'' or make up false information, making their use risky in high-stakes domains. Recently, certain legal research providers have touted methods such as retrieval-augmented generation (RAG) as ``eliminating''~\citep{casetext:eliminates} or  ``avoid[ing]'' hallucinations~\citep{westlaw:avoids}, or guaranteeing ``hallucination-free'' legal citations~\citep{lexis:hfree}.  Because of the closed nature of these systems, systematically assessing these claims is challenging. In this article, we design and report on the first preregistered empirical evaluation of AI-driven legal research tools. We demonstrate that the providers' claims are overstated. While hallucinations are reduced relative to general-purpose chatbots (GPT-4), we find that the AI research tools made by LexisNexis (Lexis+ AI) and Thomson Reuters (Westlaw AI-Assisted Research and Ask Practical Law AI) each hallucinate between 17\% and 33\% of the time. We also document substantial differences between systems in responsiveness and accuracy. 
Our article makes four key contributions. It is the first to assess and report the performance of RAG-based proprietary legal AI tools. Second, it introduces a comprehensive, preregistered dataset for identifying and understanding vulnerabilities in these systems. Third, it proposes a clear typology for differentiating between hallucinations and accurate legal responses. Last, it provides evidence to inform the responsibilities of legal professionals in supervising and verifying AI outputs, which remains a central open question for the responsible integration of AI into law.\footnote{Our dataset, tool outputs, and labels will be made available upon publication. This version of the manuscript (\today) is updated to reflect an evaluation of Westlaw's AI-Assisted Research.}
\end{abstract}

\section{Introduction}
\label{sec:introduction}

\begin{figure}[t]
    \centering
    \includegraphics{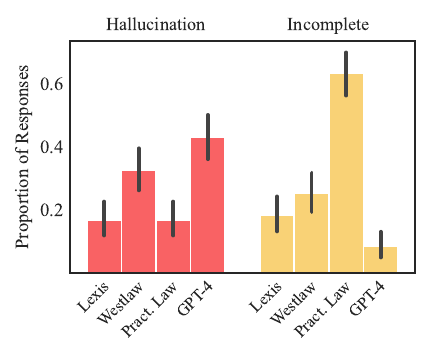}
    \caption{Comparison of hallucinated and incomplete answers across generative legal research tools. Hallucinated responses are those that include false statements or falsely assert a source supports a statement. Incomplete responses are those that fail to either address the user's query or provide proper citations for factual claims.}
    \label{fig:overall}
\end{figure}

In the legal profession, the recent integration of large language models (LLMs) into research and writing tools presents both unprecedented opportunities and significant challenges~\citep{aba}. These systems promise to perform complex legal tasks, but their adoption remains hindered by a critical flaw: their tendency to generate incorrect or misleading information, a phenomenon generally known as ``hallucination'' ~\citep{Dahl2024}.

As some lawyers have learned the hard way, hallucinations are not merely a theoretical concern \citep{Weiser2023a}. In one highly-publicized case, a New York lawyer faced sanctions for citing ChatGPT-invented fictional cases in a legal brief~\citep{Weiser2023}; many similar incidents have since been documented~\citep{Weiser2023a}. In his 2023 annual report on the judiciary, Chief Justice John Roberts specifically noted the risk of ``hallucinations'' as a barrier to the use of AI in legal practice \citep{Roberts2023}.

Recently, however, legal technology providers such as LexisNexis and Thomson Reuters (parent company of Westlaw) have claimed to mitigate, if not entirely solve, hallucination risk~\citep[][\emph{inter alia}]{lexis:hfree,casetext:eliminates,westlaw:avoids}. They say their use of sophisticated techniques such as retrieval-augmented generation (RAG) largely prevents hallucination in legal research tasks.\footnote{The following are official statements from Lexis, Casetext, and  Thomson Reuters; however, none of them has provided any clear evidence so far to support their claims about the capabilities of their AI-based legal research tools:

\vspace{0.2em}

\noindent
\textbf{Lexis:} ``Unlike other vendors, however, \emph{Lexis+ AI delivers 100\% hallucination-free linked legal citations} connected to source documents, grounding those responses in authoritative resources that can be relied upon with confidence.''~\citep[][]{Wellen2024a} (emphasis added).

\noindent
\textbf{Casetext:} ``Unlike even the most advanced LLMs, \emph{CoCounsel does not make up facts, or `hallucinate,'} because we’ve implemented controls to limit CoCounsel to answering from known, reliable data sources—such as our comprehensive, up-to-date database of case law, statutes, regulations, and codes—or not to answer at all.''~\citep{casetext:eliminates} (emphasis added). 

\noindent
\textbf{Thomson Reuters:} ``\emph{We avoid [hallucinations] by relying on the trusted content within Westlaw} and building in checks and balances that ensure our answers are grounded in good law.''~\citep{westlaw:avoids} (emphasis added).
``We’ve all heard horror stories where generative AI just makes things up. That doesn’t work for the legal industry. They have to trust the content that AI serves up. With Ask Practical Law AI, \emph{all the responses are based on the expert resources} of Practical Law.'' \citep{westlaw:practicallawai} (emphasis added)

}  (We provide details on RAG systems in Section~\ref{subsec:rag} below.) 

But none of these bold proclamations have been accompanied by empirical evidence. Moreover, the term ``hallucination'' itself is often left undefined in marketing materials, leading to confusion about which risks these tools genuinely mitigate. This study seeks to address these gaps by evaluating the performance of AI-driven legal research tools offered by LexisNexis (Lexis+ AI) and Thomson Reuters (Westlaw AI-Assisted Research and Ask Practical Law AI), and, for comparison, GPT-4.  

\begin{figure}[t]
    \centering
\includegraphics[width=\linewidth]{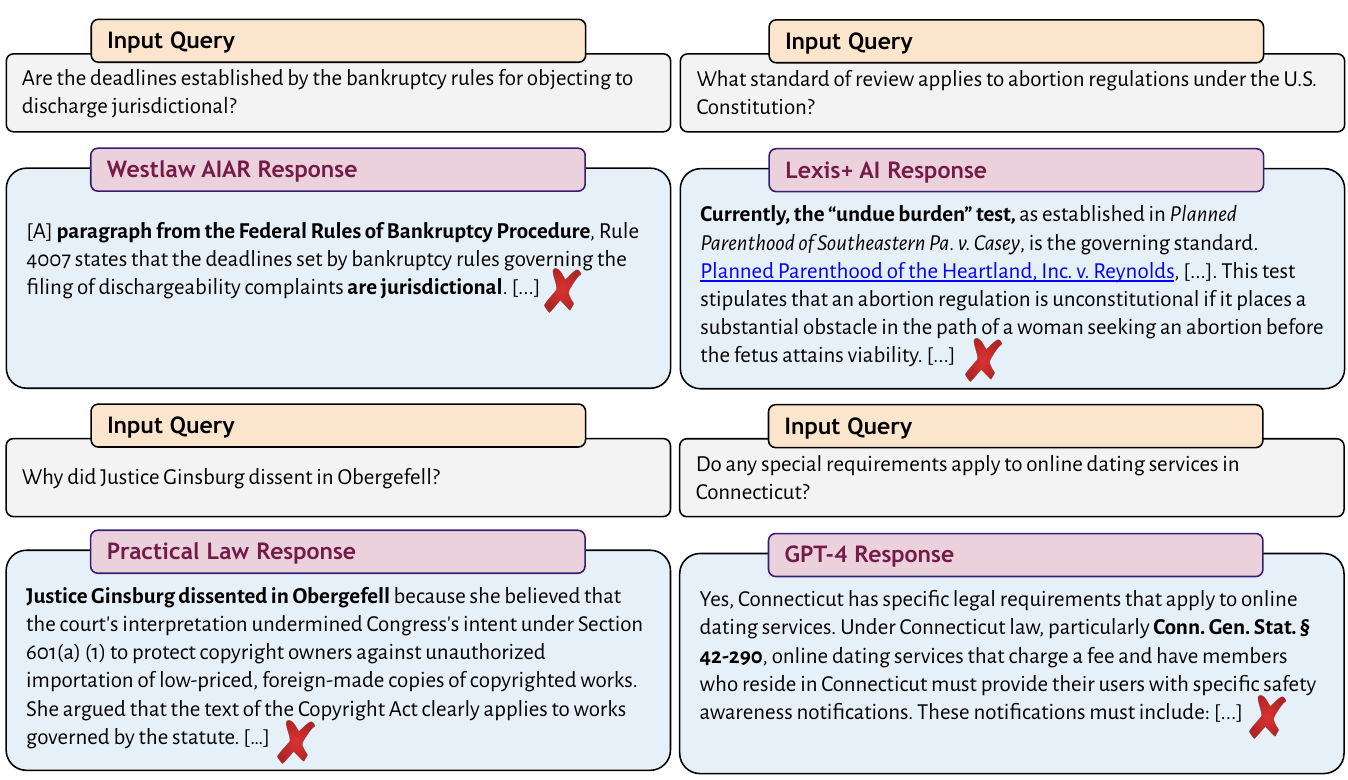}
    \caption[]{\textit{Top left:} Example of a hallucinated response by Westlaw's AI-Assisted Research product. The system makes up a statement in the Federal Rules of Bankruptcy Procedure that does not exist. \textit{Top right:} Example of a hallucinated response by LexisNexis's Lexis+ AI. \emph{Casey} and its undue burden standard were overruled by the Supreme Court in \emph{Dobbs v.\ Jackson Women's Health Organization}, 597 U.S. 215 (2022); the correct answer is rational basis review. \textit{Bottom left:} Example of a hallucinated response by Thomson Reuters's Ask Practical Law AI. The system fails to correct the user's mistaken premise---in reality, Justice Ginsburg joined the Court's landmark decision legalizing same-sex marriage---and instead provides additional false information about the case. \textit{Bottom right:} Example of a hallucinated response from GPT-4, which generates a statutory provision that does not exist.}
    \label{fig:westlaw-hallucination}
\end{figure}

Our findings, summarized in Figure~\ref{fig:overall}, reveal a more nuanced reality than the one presented by these providers: while RAG appears to improve the performance of language models in answering legal queries, the hallucination problem persists at significant levels.  To offer one simple example, shown in the top left panel of Figure~\ref{fig:westlaw-hallucination}, the Westlaw system claims that a paragraph in the Federal Rules of Bankruptcy Procedure (FRBP) states that deadlines are jurisdictional. But no such paragraph exists, and the underlying claim is itself unlikely to be true in light of the Supreme Court's holding in \emph{Kontrick v. Ryan}, 540 U.S. 443, 447-48 \& 448 n.3 (2004), which held that FRBP deadlines under a related provision were not jurisdictional.\footnote{We ran the queries for Lexis+ AI and Thomson Reuters Ask Practical Law AI in Figure~\ref{fig:westlaw-hallucination} as a test prior to the creation of our benchmark dataset; because our queries for the evaluation presented in this article were preregistered, these two examples are not included in our results below.}

We also document substantial variation in system performance. LexisNexis's Lexis+ AI is the highest-performing system we test, answering 65\% of our queries accurately. Westlaw's AI-Assisted Research is accurate 42\% of the time, but hallucinates nearly twice as often as the other legal tools we test. And Thomson Reuters's Ask Practical Law AI provides incomplete answers (refusals or ungrounded responses; see Section~\ref{sec:hallucination_definition}) on more than 60\% of our queries, the highest rate among the systems we tested.

Our article makes four key contributions. First, we conduct the first
systematic assessment of leading AI tools for real-world legal research tasks.
Second, we manually construct a preregistered dataset of over 200 legal queries
for identifying and understanding vulnerabilities in legal AI tools. We run
these queries on LexisNexis (Lexis+ AI), Thomson Reuters (Ask Practical Law AI), Westlaw (AI-Assisted Research), and GPT-4 and manually review their
outputs for accuracy and fidelity to authority. Third, we offer  a detailed typology
to refine the understanding of ``hallucinations,'' which enables us to
rigorously assess the claims made by AI service providers. Last, We not
only uncovers limitations of current technologies, but also
characterize the reasons that they fail. These results inform the
responsibilities of legal professionals in supervising and verifying AI outputs,
which remains an important open question for the responsible integration of AI
into law.

The rest of this work is organized as follows. Section~\ref{sec:background}
provides an overview of the rise of AI in law and discusses the central
challenge of hallucinations. Section~\ref{sec:rag} describes the potential and
limitations of RAG systems to reduce hallucinations. Section~\ref{sec:concepts} proposes a framework for evaluating hallucinations in a legal RAG system.
Because legal research
commonly requires the inclusion of citations, we define a \emph{hallucination}
as a response that contains either incorrect information or a false assertion that a source supports a proposition. 
Section~\ref{sec:methodology} details our methodology to evaluate the
performance of AI-based legal research tools (legal AI tools). Section~\ref{sec:results} presents our
results. We find that legal RAG can reduce hallucinations compared to
general-purpose AI systems (here, GPT-4), but hallucinations remain substantial, wide-ranging, and potentially insidious.  Section~\ref{sec:limits} discusses the limitations of our
study and the challenges of evaluating proprietary legal AI systems, which have far more restrictive conditions of use than AI systems available in other domains. Section~\ref{sec:implications} discusses the implications for legal practice and legal AI companies. 
Section~\ref{sec:conclude} concludes
with implications of our findings for legal practice. 
 
\section{Background}
\label{sec:background}
\subsection{The Rise and Risks of Legal AI}

Lawyers are increasingly using AI to augment their legal practice, and with good reason: from drafting contracts, to analyzing discovery productions, to conducting legal research, these tools promise significant efficiency gains over traditional methods. As of January 2024, at least 41 of the top 100 largest law firms in the United States have begun to use some form of AI in their practice \citep{Henry2024}; among a broader sample of 384 firms, 35\% now report working with at least one generative AI provider \citep{Collens2024}. And in a recent survey of 1,200 lawyers practicing in the United Kingdom, 14\% say that they are using generative AI tools weekly or more often \citep{Greenhill2024}.

However, adoption of these tools is not without risk. Legal AI tools present unprecedented ethical challenges for lawyers, including concerns about client confidentiality, data protection, the introduction of new forms of bias, and lawyers’ ultimate duty of supervision over their work product \citep{Avery2023, Cyphert2021, Walters2019, Yamane2020}. Recognizing this, the bar associations of California \citeyearpar{ca-bar2023}, New York \citeyearpar{ny-bar2024}, and Florida \citeyearpar{florida-bar2024} have all recently published guidance on how AI should be safely and ethically integrated into their members’ legal practices. Courts have weighed in as well: as of May 2024, more than 25 federal judges have issued standing orders instructing attorneys to disclose or limit the use of AI in their courtrooms \citep{Law3602024}.

In order for these guidelines to be effective, however, lawyers need to first understand what exactly an AI tool is, how it works, and the ways in which it might expose them to liability. Do different tools have different error rates---and what kinds of errors are likely to manifest? What training do lawyers need in order to spot these errors---and can they do anything as users to mitigate them? Are there particular tasks that current AI tools are particularly adept at---and are there any that lawyers should stay away from?

This paper moves beyond previous work on general-purpose AI tools \citep{Choi2024, Dahl2024, Schwarcz2023} by answering these questions specifically for \textit{legal} AI tools---namely, the tools that have been carefully developed by leading legal technology companies and that are currently being marketed to lawyers as avoiding many of the risks known to exist in off-the-shelf offerings. In doing so, we aim to provide the concrete empirical information that lawyers need in order to assess the ethical and practical dangers of relying on these new commercial AI products.

\subsection{The Hallucination Problem}

We focus on one problem of AI that has received considerable attention in the legal community: ``hallucination,'' or the tendency of AI tools to produce outputs that are demonstrably false.\footnote{Theoretical work has shown that hallucinations must occur at a certain rate for calibrated generative language models, regardless of their architecture, training data quality, or size \citep{Kalai2023}.}  In multiple high-profile cases, lawyers have been reprimanded for submitting filings to courts citing nonexistent case law hallucinated by an AI service \citep{Weiser2023, Weiser2023a}.  Previous work has found that general-purpose LLMs hallucinate on legal queries on average between 58\% and 82\% of the time \citep{Dahl2024}. Yet this prior work did not examine tools specifically developed for the legal setting, such as tools that use LLMs with auxiliary legal databases and RAG. And because these tools are placed prominently before lawyers on leading legal research platforms (i.e., LexisNexis and Thomson Reuters / Westlaw), a systematic examination is sorely needed. 

In this article, we focus on \textit{factual} hallucinations. In the legal setting, there are three primary ways that a model can be said to hallucinate: it can be unfaithful to its training data,
unfaithful to its prompt input, or unfaithful to the true facts of the world
\citep{Dahl2024}. Because we are interested in legal research tools that are meant to help lawyers understand legal facts, we focus on the third category: factual hallucinations.\footnote{Other definitions of hallucination could be more relevant in other contexts. For example, future research should examine AI tools for contract analysis or document summarization. For that analysis, it would be more important to study hallucinations with respect to the tool's input prompt, rather than with respect to the general facts of the world. Evaluation standards for such generative AI output, however, are still in flux.} However, in Section~\ref{sec:hallucination_definition} below, we also expand on this definition by decomposing factual hallucinations into two dimensions: \textit{correctness} and \textit{groundedness}. We hope that this distinction will provide useful guidance for users seeking to understand the precise way that these tools can be helpful or harmful. 

\section{Retrieval-Augmented Generation (RAG)}
\label{sec:rag}

\begin{figure}[t]
    \centering
\includegraphics[width=0.97\linewidth]{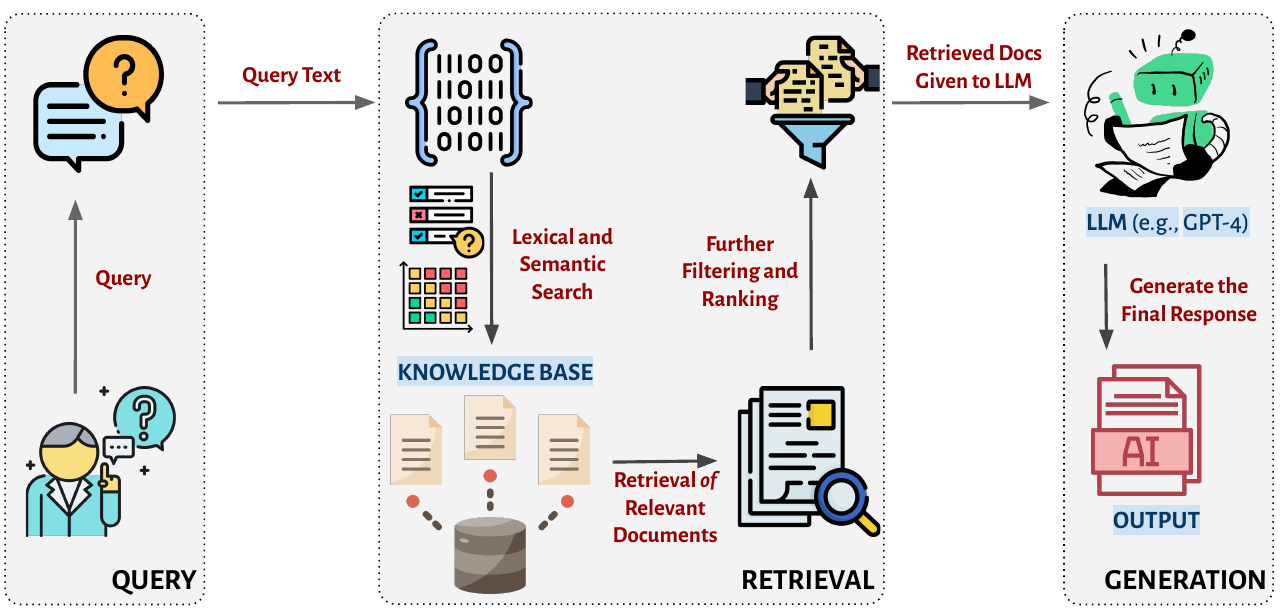}
    \caption[]{Schematic diagram of a retrieval-augmented generation (RAG) system. Given a user query (left), the typical process consists of two steps: (1) retrieval (middle), where the query is embedded with natural language processing and a retrieval system takes embeddings and retrieves the relevant documents (e.g., Supreme Court cases); and (2) generation (right), where the retrieved texts are fed to the language model to generate the response to the user query. Any of the subsidiary steps may introduce error and hallucinations into the generated response. (Icons are credited to FlatIcon.)}
    \label{fig:RAG}
\end{figure}

\subsection{The Promise of RAG}
\label{subsec:rag}

Across many domains, the fairly new technique of retrieval-augmented generation (RAG) is being seen and heavily promoted as the key technology for making LLMs effective in domain-specific contexts. It allows general LLMs to make effective use of company- or domain-specific data and to produce more detailed and accurate answers by drawing directly from retrieved text. In particular, 
RAG is commonly touted as the solution for legal hallucinations. In a February 2024 interview, a Thomson Reuters executive asserted that, within Westlaw AI-Assisted Research, RAG ``dramatically reduces hallucinations to nearly zero'' \citep{Ambrogi2024}. Similarly, LexisNexis has said that RAG enables it to ``deliver accurate and authoritative answers that are grounded in the closed universe of authoritative content'' \citep{Wellen2024}.\footnote{In Section~\ref{sec:hallucination_definition} below, we discuss how different companies may be using definitions of ``hallucination'' different from the ones more commonly accepted in the literature or in popular discourse.}

As depicted in Figure~\ref{fig:RAG}, RAG comprises two primary steps to transform a query into a response: (1) retrieval and (2) generation \citep{Lewis2020, Gao2024}. Retrieval is the process of selecting relevant documents from a large universe of documents. This process is familiar to anyone who uses a search engine: using keywords, user information, and other context, a search engine quickly identifies a handful of relevant web pages out of the millions available on the internet. Retrieval systems can be simple, like a keyword search, or complex, involving machine learning techniques to capture the semantic meaning of a query (such as neural text embeddings).

With the retrieved documents in hand, the second step of generation involves providing those documents to a LLM along with the text of the original query, allowing the LLM to use \textit{both} to generate a response. Many RAG systems involve additional pre- and post-processing of their inputs and outputs (e.g., filtering and extraction depicted in the middle panel of Figure~\ref{fig:RAG}), but retrieval and generation are the hallmarks of a RAG pipeline.

The advantage of RAG is obvious: including retrieved information in the prompt allows the model to respond in an ``open-book'' setting rather than in ``closed-book'' one. The LLM can use the information in the retrieved documents to inform its response, rather than its hazy internal knowledge. Instead of generating text that conforms to the general trends of a highly compressed representation of its training data, the LLM can rely on the full text of the relevant information that is injected directly into its prompt.

For example, suppose that an LLM is asked to state the year that \textit{Brown v.\ Board of Education} was decided. In a closed-book setting, the LLM, without access to an external knowledge base, would generate an answer purely based on its internal knowledge learned during training---but a more obscure case might have little or no information present in the training data, and the model could generate a realistic-sounding year that may or may not be accurate. In a RAG system, by contrast, the retriever would first look up the case name in a legal database, retrieve the relevant metadata, and then provide that to the LLM, which would use the result to provide the user a response to their query.

On paper, RAG has the potential to substantially mitigate many of the kinds of legal hallucinations that are known to afflict off-the-shelf LLMs \citep{Dahl2024}---the technique performs well in many general question-answering situations \citep{Guu2020, Lewis2020, Siriwardhana2023}. However, as we show in the next section, RAG systems are no panacea.
 
\subsection{Limitations of RAG}
\label{sec:rag_limitations} 

There are several reasons that RAG is unlikely to fully solve the hallucination problem \citep{Barnett2024}. Here, we highlight some that are unique to the legal domain.

First, retrieval is particularly challenging in law. Many popular LLM benchmarking datasets \citep{Rajpurkar2016, Yang2018} contain questions with clear, unambiguous references that address the question in the source database. Legal queries, however, often do not admit a single, clear-cut answer \citep{Mik2024}. In a common law system, case law is created over time by judges writing opinions; this precedent then builds on precedent in the way that a chain novel might be written in \textit{seriatim} \citep{Dworkin1986}. By construction, these legal opinions are not atomic \textit{facts}; indeed, on some views, the law is an ``essentially contested'' concept \citep{Waldron2002}. Thus, deciding what to retrieve can be challenging in a legal setting. At best, a RAG system must be able to locate information from multiple sources across time and place in order to properly answer a query. And at worst, there may be no set of available documents that definitively answers the query, if the question presented is novel or indeterminate.

Second, document relevance in the legal context is not based on text alone. Most retrieval systems identify relevant documents based on some kind of text similarity \citep{Karpukhin2020}. But the retrieval of documents that only \textit{seem} textually relevant---and are ultimately irrelevant, or ``distracting''---negatively affects performance on general question-answering tasks \citep{Cuconasu2024, Chen2024}. Problems of this type are likely to compound in the legal domain. In different jurisdictions and in different time periods, the applicable rule or the relevant jurisprudence may differ. Even similar-sounding text in the correct time and place may not apply if special conditions are not met. The problem may be worse if a rule that applies in a special condition conflicts with a more broadly applicable rule. The LLM may have been trained on a much greater volume of text supporting the broadly applicable rule, and may be more faithful to its training data than to the retrieval context. Consequently, designing a high-quality research tool that deals with this problem requires careful attention to non-textual elements of retrieval and the deference of the model to different sources of information. These challenges are familiar to lawyers using ``natural language'' searches, as opposed to more deterministic boolean searches, on legal research platforms. 

Third, the generation of meaningful legal text is also far from straightforward. Legal documents are generally written for other lawyers immersed in the same issue, and they rely on an immense amount of background knowledge to properly understand and apply. A helpful generative legal research tool would have to do far more than simple document summarization; it would need to synthesize facts, holdings, and rules from different pieces of text while keeping the appropriate legal context in mind. For example, consider this Lexis+ AI exchange:
\begin{quote}
    \textbf{Prompt:} What are some notable opinions written by Judge Luther A. Wilgarten?

    \textbf{Lexis+ AI Response:} One of the notable opinions written by Judge
    Luther A. Wilgarten is the case of Luther v.\ Locke. The case citation is
    2010 U.S. Dist. LEXIS 144868 and it was decided in the Eastern District
    of Virginia on January 11, 2010. In this case, the plaintiff had filed a
    Notice of Appeal of Judge Ellis's decision, but failed to properly
    prosecute the appeal. [...]
\end{quote}

While the retrieved citation offered is a real case and hence ``hallucination-free'' in a narrow sense, it was not written by Judge Wilgarten, a fictional judge who never served on the bench \citep{Miner1989}.\footnote{This retrieval error likely reflects the similarity in the embedding space between ``Judge Luther A. Wilgarten'' and the terms ``judge'' (mentioned 9 times in the 900-some word order) and ``William Luther,'' the plaintiff in the case.} And while the generated passages are based on the actual case, the second sentence  contradicts the premise, suggesting Judge \emph{Ellis} wrote the opinion, but the opinion was actually written by Judge Brinkema (and involved a prior decision by Judge Ellis, which forms the basis for the RAG response). Nor is the decision notable, as it was an unpublished opinion cited only once outside of its direct history. Hallucinations are compounded by poor retrieval and erroneous generation. 

Conceptualizing the potential failure modes of legal RAG systems requires domain expertise in both computer science \textit{and} law. As is apparent once we examine the component parts of a RAG system in Figure~\ref{fig:RAG}, each of the subsidiary steps (the embedding, the design of lexical and semantic search, the number of documents retrieved, and filtering and extraction) involves design choices that can affect the quality of output \citep{Barnett2024}, each with potentially subtle trade-offs \citep{belkin2008some}.  In the next section, we devise a new task suite specifically designed to probe the prevalence of RAG-resistant hallucinations, complementing existing benchmarking efforts that target AI's legal knowledge in general \citep{Dahl2024} and its capacity for legal reasoning \citep{Guha2023}.

\section{Conceptualizing Legal Hallucinations}
\label{sec:concepts}

The binary notion of hallucination developed in \citet{Dahl2024} does not fully capture the behavior of RAG systems, which are intended to generate information that is both accurate and grounded in retrieved documents. We expand the framework of legal hallucinations to \textit{two} primary dimensions: correctness and groundedness. Correctness refers to the factual accuracy of the tool's response (Section~\ref{sec:correctness_definition}). Groundedness refers to the relationship between the model's response and its cited sources (Section~\ref{sec:groundedness_definition}).

Decomposing factual hallucinations in this way enables a more nuanced analysis and 
understanding of how exactly legal AI tools fail in practice. For example, a
response could be correct but improperly grounded. This might happen when
retrieval results are poor or irrelevant, but the model happens to produce the correct answer, falsely asserting that an unrelated source supports its conclusion. This can mislead the user in potentially dangerous ways.

\begin{table*}
\begin{tabular}{p{2.0cm}p{4.7cm}p{6cm}}
\toprule
\textbf{} & \textbf{Description} & \textbf{Example} \\
\midrule
\multicolumn{3}{l}{\textit{Correctness}} \\
\cmidrule{1-3}
Correct & Response is factually correct and relevant & The right to same sex marriage is protected under the U.S. Constitution. \emph{Obergfell v.\ Hodges}, 576 U.S. 644 (2015). \\
\hlred{Incorrect} & Response contains factually \mbox{inaccurate} information & There is no right to same sex marriage in the United States. \\
Refusal & Model refuses to provide any answer or provides an irrelevant answer & I'm sorry, but I cannot answer that question. Please try a different query. \\
\midrule
\multicolumn{3}{l}{\textit{Groundedness}} \\
\cmidrule{1-3}
Grounded & Key factual propositions make valid references to relevant legal documents & The right to same sex marriage is protected under the U.S. Constitution. \emph{Obergfell v.\ Hodges}, 576 U.S. 644 (2015). \\
\hlred{Misgrounded} & Key factual propositions are cited but the source does not support the claim & The right to same sex marriage is protected under the U.S. Constitution. \emph{Miranda v.\ Arizona}, 384 U.S. 436 (1966). \\
Ungrounded & Key factual propositions are not cited & The right to same sex marriage is protected under the U.S. Constitution. \\
\bottomrule
\end{tabular}
\caption{\label{tab:coding-summary}A summary of our coding criteria for correctness and groundedness, along with hypothetical responses to the query ``Does the Constitution protect a right to same sex marriage?'' that would fall under each of the categories. Groundedness is only applicable for correct responses. The categories which qualify as a ``hallucination'' are highlighted in \hlred{red}.}
\end{table*}

\subsection{Correctness}
\label{sec:correctness_definition}

We say that a response is \emph{correct} if it is both factually correct and relevant to the query. A response is \emph{incorrect} if it contains any factually inaccurate information. For the purposes of this analysis, we label an answer that is partially correct---that is, one that contains correct information that does not fully address the question---as correct. If a response is neither correct nor incorrect, because the model simply declines to respond, we label that as a \textit{refusal}. See the top panel of Table~\ref{tab:coding-summary} for examples of each of these three codings of correctness.\footnote{Note that for our false premise questions, the desired behavior is for the model to refute and state the false assumption in the user's prompt. A gold-standard response to such a question would therefore be a statement that the assumption may be incorrect, with a case law citation to the opposite proposition. However, for these false premise questions alone, we also label a refusal which mentions the fact that no pertinent sources were found as correct.}

\subsection{Groundedness}
\label{sec:groundedness_definition}

For correct responses, we additionally evaluate each response's groundedness. A response is \emph{grounded} if the key factual propositions in its response make valid references to relevant legal documents. A response is \emph{ungrounded} if key factual propositions are not cited. A response is \emph{misgrounded} if key factual propositions  are cited but misinterpret the source or reference an inapplicable source. See the bottom panel of Table~\ref{tab:coding-summary} for examples illustrating groundedness.

Note that our use of the term \emph{grounded} deviates somewhat from the notion in computer science. In the computer science literature, groundedness refers to adherence to the source documents provided, regardless of the relevance or accuracy of the provided documents \citep{Agrawal2023}. In this paper, by contrast, we evaluate the quality of the retrieval system and the generation model together in the legal context. Therefore, when we say \emph{grounded}, we mean it in the legal sense---that is, responses that are correctly grounded in actual governing caselaw. If the retrieval system provides documents that are inappropriate to the jurisdiction of interest, and the model cites them in its response, we call that \emph{misgrounded}, even though this might be a technically ``grounded'' response in the computer-science sense.

\subsection{Hallucination}
\label{sec:hallucination_definition}

We now adopt a precise definition of a hallucination in terms of the above variables. A response is considered \emph{hallucinated} if it is either incorrect or misgrounded. In other words, if a model makes a false statement or falsely asserts that a source supports a statement, that constitutes a hallucination.

This definition provides technical clarity to the popular concept of hallucination, which is a term that is currently being used inconsistently by different industry actors. For example, in one interview, one Thomson Reuters executive appeared to refer to hallucinations as exclusively instances when an AI system fabricates the \emph{existence} of a case, statute, or regulation, distinct from more general problems of accuracy \citep{Ambrogi2024}. Yet, in a December 2023 press release, another Thomson Reuters executive defined hallucinations differently, as ``responses that sound plausible but are completely false'' \citep{westlaw:avoids}.

LexisNexis, by contrast, uses the term hallucination in yet a different way. LexisNexis claims that its AI tool provides ``linked hallucination-free legal citations'' \citep{lexis:hfree}, but, as we demonstrate below, this claim can only be true in the most narrow sense of ``hallucination,'' in that their tool does indeed \textit{link} to real legal documents.\footnote{Of course, there is some evidence that Lexis+ AI does not succeed even by this metric. \citet{tweet} reports instances of Lexis+ AI citing cases decided in 2025.} If those linked sources are irrelevant, or even contradict the AI tool's claims, the tool has, in our sense, engaged in a hallucination. Failing to capture that dimension of hallucination would require us to conclude that a tool that links only to \emph{Brown v.\ Board of Education} on every query (or provides cases for fictional judges as in the instance of Luther A.\ Wilgarten) has provided ``hallucination-free'' citations, a plainly irrational result.

More concretely, consider the \emph{Casey} example in Figure~\ref{fig:westlaw-hallucination}, where the linked citation \emph{Planned Parenthood v.\ Reynolds} is a real case that has not been overturned.\footnote{\emph{Reynolds} even appears in the citation list with a positive Shepardization symbol.} However, the model's answer relies on \emph{Reynolds}' description of \emph{Planned Parenthood v.\ Casey}, a case that has been overturned. The model's response is incorrect, and its citation serves only to mislead the user about the reliability of its answer \citep{Goddard2012}.

These errors are potentially more dangerous than fabricating a case outright, because they are subtler and more difficult to spot.\footnote{As \citet{Gottlieb2024} reports in one the assessment by law firms of generative AI products, ``The importance of reviewing and verifying the accuracy of the output, including checking the AI’s answers against other sources, makes any efficiency gains difficult to measure.''} Checking for these kinds of hallucinations requires users to click through to cited references, read and understand the relevant sources, assess their authority, and compare them to the propositions the model seeks to support. Our definition reflects this more complete understanding of ``hallucination.''

\subsection{Accuracy and Incompleteness}

Alongside \emph{hallucinations}, we also define two other top-level labels in terms of our correctness and groundedness variables: \emph{accurate responses}, which are those that are both correct and grounded, and \emph{incomplete responses}, which are those that are either refusals or ungrounded.

We code correct but ungrounded responses as incomplete because, unlike a
misgrounded response, an  ungrounded response does not actually make any false
assertions. Because an ungrounded
response does not provide key information (supporting authorities) that the user
needs, it is marked incomplete.

\section{Methodology}
\label{sec:methodology}

\subsection{AI-Driven Legal Research Tools}

We study the hallucination rate and response quality of three available RAG-based AI research tools: LexisNexis's Lexis+ AI, Thomson Reuters's Ask Practical Law AI, and Westlaw's AI-Assisted Research. As nearly every practicing U.S.\ lawyer knows, Thomson Reuters (the parent company of Westlaw) and LexisNexis\footnote{LexisNexis is owned by the RELX Group.} have historically enjoyed a virtual duopoly over the legal research market \citep{Arewa2006} and continue to be two of the largest incumbents now selling legal AI products \citep{Ma2024}.

\textbf{Lexis+ AI} functions as a standard chatbot interface, like ChatGPT, with a text area for the user to enter an open-ended inquiry. In contrast to traditional forms of legal search, ``boolean'' connectors and search functions like \texttt{AND}, \texttt{OR}, and \texttt{W/n} are neither required nor supported. Instead, the user simply formulates their query in natural language, and the model responds in kind. The user then has the option to continue the chat by asking another question, which the tool will respond to with the complete context of both questions. Introduced in October 2023, Lexis+ AI states that it has access to LexisNexis's entire repository of case law, codes, rules, constitution, agency decisions, treatises, and practical guidance, all of which it presumably uses to craft its responses. While not much technical detail is published, it is known that Lexis+ AI implements a proprietary RAG system that ensures that every prompt ``undergoes a minimum of five crucial checkpoints \dots to produce the highest quality answer'' \citep{Wellen2024a}.\footnote{Since the completion of our evaluation for this paper in April 2024, LexisNexis has released a ``second generation'' version of its tool. Our results do not speak to the performance of this second generation product, if different. Accompanying this release, LexisNexis noted, ``our promise is not perfection, but that all linked legal citations are hallucination-free'' \citep{LexisNexis2024}.}

\textbf{Ask Practical Law AI}, introduced in January 2024 and offered on the Westlaw platform, is a more limited product, but it operates in a similar way. Like Lexis+ AI, Ask Practical Law AI also functions as a chatbot, allowing the user to input their queries in natural language and responding to them in the same format. However, instead of accessing all the primary sources that Lexis+ AI uses, Ask Practical Law AI only retrieves information from Thomson Reuters's database of ``practical law'' documents---``expert resources \dots that have been created and curated by more than 650 bar-admitted attorney editors'' \citep{westlaw:practicallawai} promising ``90,000+ total resources across 17 practice areas'' \citep{PracticalLaw}. Thomson Reuters markets this database for general legal research: ``Practical Law provides trusted, up-to-date legal know-how across all major practice areas to help attorneys deliver accurate answers quickly and confidently.'' Performing RAG on these materials, Thomson Reuters claims, ensures that its system ``only returns information from [this] universe'' \citep{westlaw:practicallawai}.

\textbf{Westlaw's AI-Assisted Research (AI-AR)}, introduced in November 2023, is also a standard chatbot interface, promising ``answers to a far broader array of questions than what we could anticipate with human power alone'' \citep{westlaw:avoids}. The RAG system retrieves information from Westlaw's databases of cases, statutes, regulations, West Key Numbers, headnotes, and KeyCite markers \citep{westlaw:avoids}. While not much technical detail is provided, AI-AR appears to rely on OpenAI's GPT-4 system \citep{Ambrogi2023}. This system was built out after a \$650 million acquisition of Casetext, which had developed legal research systems on top of GPT-4 \citep{Ambrogi2023}. RAG is prominently touted as addressing hallucinations: one Thomson Reuters official stated, ``We avoid [hallucinations] by relying on the trusted content within Westlaw and building in checks and balances that ensure our answers are grounded in good law'' \citep{westlaw:avoids}. While AI-AR has been sold to law firms, it has not been been made available generally for educational and research purposes.\footnote{Thomson Reuters denied three requests for access by our team at the time we conducted our initial evaluation. The company provided access after the initial release of our results.} 

Both AI-AR and Ask Practical Law AI are made available via the Westlaw platform and are commonly referred to as AI products within Westlaw.\footnote{The home page of Practical Law is titled ``Practical Law US - Westlaw'' and is located on a subdomain of \texttt{westlaw.com} \citep{wayback:google_practical_law}. See also, e.g., \citet{berkeley} (noting that ``Ask Practical Law AI'' is now available on Westlaw''); 
\citet{vanderheijden:ai_products} (describing ``Ask Practical Law AI'' as a Westlaw product); \citet{uwlibrary} (describing ``Practic[al] Law [a]s a database within Westlaw''); \citet{moakleylaw:practical_ai} (noting ``Ask Practical Law AI (Westlaw)''); \citet{ellie} (writing that ``Westlaw released Ask Practical Law AI to academic accounts'').} For shorthand, we will refer to Ask Practical Law AI as a Thomson Reuters system and AI-AR as a Westlaw system, as this appears to track the internal company product distinctions.%

To provide a point of reference for the quality of these bespoke legal research tools---and because AI-AR appears to be built on top of GPT-4---we also evaluate the hallucination rate and response quality of GPT-4, a widely available LLM that has been adopted as a knowledge-work assistant \citep{dellacqua2023, Collens2024}. GPT-4's responses are produced in a ``closed-book'' setting; that is, produced without access to an external knowledge base.

\begin{table*}
\begin{tabular}{p{1.5cm}p{1cm}p{1cm}p{3.5cm}p{5.3cm}}
\toprule
\textbf{Category} & \textbf{Count} & \textbf{Perc.} & \textbf{Description} & \textbf{Example Query} \\
\midrule
General legal research & 80 & 39.6\% & Common-law doctrine questions, previously published practice bar exam questions, holding questions & Has a habeas petitioner's claim been ``adjudicated on the merits'' for purposes of 28 U.S.C. § 2254(d) where the state court denied relief in an explained decision but did not expressly acknowledge a federal-law basis for the claim? \\
\cmidrule{1-5}
Jurisdiction or time-specific & 70 & 34.7\% & Questions about circuit splits, overturned cases, or new developments & In the Sixth Circuit, does the Americans with Disabilities Act require employers to accommodate an employee's disability that creates difficulties commuting to work? \\
\cmidrule{1-5}
False premise & 22 & 10.9\% & Questions where the user has a mistaken understanding of the law & I'm looking for a case that stands for the proposition that a pedestrian can be charged with theft for absorbing sunlight that would otherwise fall on solar panels, thereby depriving the owner of the panels of potential energy. \\
\cmidrule{1-5}
Factual recall questions & 30 & 14.9\% &  Basic queries about facts not requiring interpretation, like the year a case was decided. & Who wrote the majority opinion in Candela Laser Corp. v.\ Cynosure, Inc., 862 F. Supp. 632 (D. Mass. 1994)? \\
\bottomrule
\end{tabular}
\caption{\label{tab:queries}The high-level categories of the query dataset, with counts and percentages (Perc.) of queries, descriptions, and sample queries.}
\end{table*}
 
\subsection{Query Construction}
\label{sec:query_construction}

We design a diverse set of legal queries to probe different aspects of a legal RAG system's performance. We develop this benchmark dataset to represent real-life legal research scenarios, without prior knowledge of whether they would succeed or fail.

For ease of interpretation, we group our queries into four broad categories:
    	\begin{enumerate}
    		\item \textbf{General legal research questions:} common-law
    			doctrine questions, holding questions, or bar exam questions
    
    		\item \textbf{Jurisdiction or time-specific questions:} questions about circuit
    			splits, overturned cases, or new developments
    
    		\item \textbf{False premise questions:} questions where the user has a mistaken
    			understanding of the law
       
            \item \textbf{Factual recall questions:} queries about facts of cases not requiring interpretation, such as the author of an opinion, and matters of legal citation
    	\end{enumerate}

Queries in the first category ($n=80$) are the paradigmatic use case for these tools, asking general questions of law. For instance, such queries pose bar exam questions that have ground-truth answers, but in contrast to assessments that focus only on the accuracy of the multiple choice answer \citep[e.g.,][]{martinez2024re}, we assess hallucinations in the fully generated response. Queries in the second category ($n=70$) probe for jurisdictional differences or developing areas in the law, which represent precisely the kinds of active legal questions requiring up-to-date legal research.  Queries in the third category ($n=22$) probe for the tendency of LLMs to assume that premises in the query are true, even when flatly false. %
The last category ($n=30$) probes the extent to which RAG systems are able to overcome known vulnerabilities about how general LLMs encode legal knowledge \citep{Dahl2024}. 

Table \ref{tab:queries} describes these categories in more depth and provides an example of a question that falls within each category.  We used 20 queries from LegalBench's Rule QA task verbatim \citep{Guha2023}, and 20 BARBRI bar exam prep questions verbatim \citep{barbri}. 
Each of the 162 other queries were hand-written or adapted for use in our benchmark.
Appendix \ref{appendix:queries} provides a more granular list of the types of queries and descriptive information.

Our dataset advances AI benchmarking in five respects. First, it is expressly designed to move the evaluation of AI systems from standard question-answer settings with a discrete and known answer (e.g., multiple choice) to the generative (e.g., open-ended) setting \citep{raji2021ai, li2024task, mcintosh2024inadequacies}. Prior work has evaluated the amount of legal information that LLMs can produce \citep{Dahl2024}, but this kind of benchmark does not capture the practical benefits and risks of everyday use cases. Legal practice is more than answering multiple choice questions. Of course, because these are not simple queries, their design and evaluation is time-intensive---all queries must be written based on external legal knowledge and submitted by hand through the providers' web interfaces, and evaluation of answers requires careful assessment of the tool's legal analysis and citations, which can be voluminous.

Second, our queries are specifically tailored to RAG-based, open-ended legal research tools. This differentiates our dataset from previously released legal benchmarks, like LegalBench \citep{Guha2023}. Most LegalBench tasks are tailored towards legal analysis of information given to the model in the prompt; tasks like contract analysis or issue spotting. Our queries are written specifically for RAG-based legal research tools; each query is an open-ended legal question that requires legal analysis supported by relevant legal documents that the model must retrieve. This provides a more realistic representation of the way that lawyers are intended to use these tools. Our goal with our dataset is to move beyond anecdotal accounts and offer a systematic investigation of the potential strengths and weaknesses of these tools, responding to documented challenges in evaluating AI in law \citep{Kapoor2024, Guha2023}.

Third, these queries are designed to represent the temporal and jurisdictional variation (e.g., overruled precedents, circuit splits) that is often the subject of live legal research \citep{Beim2019}. We hypothesize that AI systems are not able to encode this type of multifaceted and dynamic knowledge at the moment, but these are precisely the kinds of inquiries requiring legal research. Due to the nature of legal authority, attorneys will inevitably have questions specific to their time, place, and facts, and even the most experienced lawyers will need to ground their understanding of the legal landscape when facing issues of first impression.

Fourth, the queries probe for ``contrafactual bias,'' or the tendency of chat systems to assume the veracity of a premise even when false \citep{Dahl2024}. Many claim that AI systems will help to address longstanding access to justice issues \citep{Bommasani2022, Chien2024, Chien2024a, Perlman2023, Tan2023}, but contrafactual bias poses particular risk for \textit{pro se} litigants and lay parties.

Last, to guard against selection bias in our results (i.e., choosing queries based on hallucination results), we modeled best practices with our dataset by preregistering our study and associated queries with the Open Science Foundation prior to performing our evaluation \citep{osf}.\footnote{We did not run any preregistered query against any tool prior to registration, with one exception, \texttt{changes-in-law-73} (``When does the undue burden standard apply in abortion cases?''). Some queries were slightly rephrased during evaluation to better elicit an answer with factual content (a prospect explicitly contemplated by the pre-registration); those queries are marked as such in our released dataset and documented in Appendix~\ref{appendix:modified_queries}.}

\subsection{Query Execution}

For Lexis+ AI, Thomson Reuters's Ask Practical Law AI, and Westlaw's AI-AR, we executed each query by copying and pasting it into the chat window of each product.\footnote{We created a new ``conversation'' for each query.} For GPT-4, we prompted the LLM via the OpenAI API (model \texttt{gpt-4-turbo-2024-04-09}) with the following instruction, appending the query afterwards:
        
    \begin{quote}
        You are a helpful assistant that answers legal questions. Do not hedge unless absolutely necessary, and be sure to answer questions precisely and cite caselaw for propositions.
    \end{quote}

This prompt aims to ensure comparability with legal AI tools, particularly by prompting for legal citations and concrete factual assertions. We recorded the complete response that each tool gave, along with any references to case law or documents.
The dataset was preregistered on March 22, 2024 and all queries on Lexis+ AI, Ask Practical Law AI, and GPT-4 were run between March 22 and April 22, 2024. Queries on Westlaw's AI-AR system were run between May 23--27, 2024.
        
\subsection{Inter-Rater Reliability}

To code each response according to the concepts of correctness, groundedness, and hallucination, we relied on our expert domain knowledge in law to hand-score each model response according to the rubric developed in Section~\ref{sec:concepts}. As noted above, efficiently evaluating AI-generated text remains an unsolved problem with inevitable trade-offs between internal validity, external validity, replicability, and speed \citep{Liu2016, Hashimoto2019, Smith2022}. These problems are particularly pronounced in our legal setting, where our queries represent real legal tasks. Accordingly, techniques of letting these legal AI tools ``check themselves''---which have become popular in other AI evaluation pipelines \citep{Manakul2023, Mundler2023, Zheng2023}---are not suitable for this application. Precisely because adherence to authority is so important in legal writing and research, our tasks must be qualitatively evaluated by hand according to the definitions of correctness and groundedness that we have carefully constructed. This makes studying these legal AI tools expensive and time-consuming: this is a cost that must be reflected in future conversations about how to responsibly integrate these AI products into legal workflows.

To ensure that our queries are sufficiently well-defined and that our coding definitions are sufficiently precise, we evaluated the inter-rater reliability of different labelers on our data. Task responses were first graded by one of three different labelers. A fourth labeler then labeled a random sample of 48 responses, stratified by model and task type. We oversampled \emph{The Bluebook} citation task slightly because it is particularly technical. The fourth labeler did not discuss anything with the first three labelers and did not have access to the initial labels. Their knowledge of the labeling process came only from our written documentation of labeling criteria, fully described in Appendix \ref{appendix:labeling}.

With this protocol, we find a Cohen's kappa \citep{Cohen1960} of 0.77 and an inter-rater agreement of 85.4\% on the final outcome label (correct, incomplete, or hallucinated) between the evaluation labeler and the initial labels. This is a substantial degree of agreement that suggests that our task and taxonomy of labels are well defined. Our results are comparable to similar evaluations for complex, hand-graded legal tasks \citep{Dahl2024}.\footnote{In updating results to include AI-AR, we also conducted another round of validation of every hallucination coding. This validation led to nearly identical results---for instance, the accuracy rate of Ask Practical Law AI in Figure~\ref{fig:overall} increased from 19\% to 20\%, which is of course within the bounds of inter-rater reliability.}

\section{Results}
\label{sec:results}

Section~\ref{sec:lexis_vs_westlaw_results} describes our findings on hallucinations and responsiveness. Section~\ref{sec:aiar} examines the varied and sometimes insidious nature of hallucinations. Section~\ref{sec:failure} provides a typology of the potential causes of inaccuracies we encountered.

    \subsection{Hallucinations Persist Across Query Types}
    \label{sec:lexis_vs_westlaw_results}

    \begin{figure}[t]
        \centering
        \includegraphics[width=\textwidth]{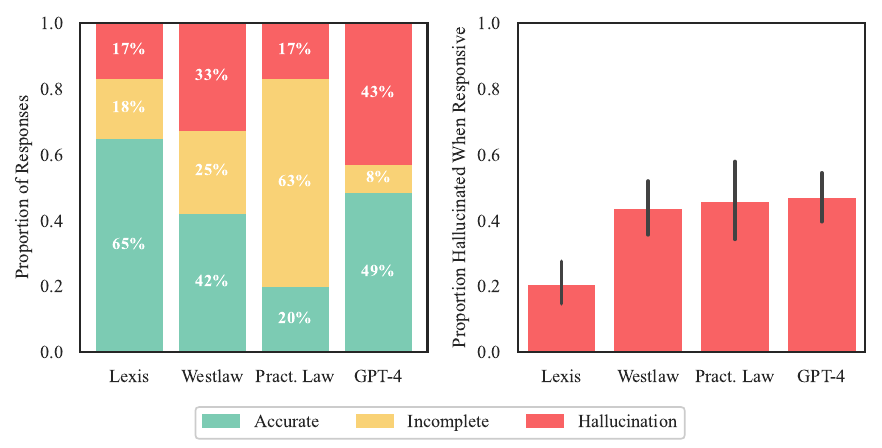}
        \caption{\emph{Left panel}: overall percentages of accurate, incomplete, and hallucinated responses. \emph{Right panel}: the percentage of answers that are hallucinated when a direct response is given. Westlaw AI-AR and Ask Practical Law AI respond to fewer queries than GPT-4, but the responses that they do produce are not significantly more trustworthy. Vertical bars denote 95\% confidence intervals.}
        \label{fig:stacked_responsive}
    \end{figure}

    \begin{figure*}[ht]
        \centering
        \includegraphics[width=\textwidth]{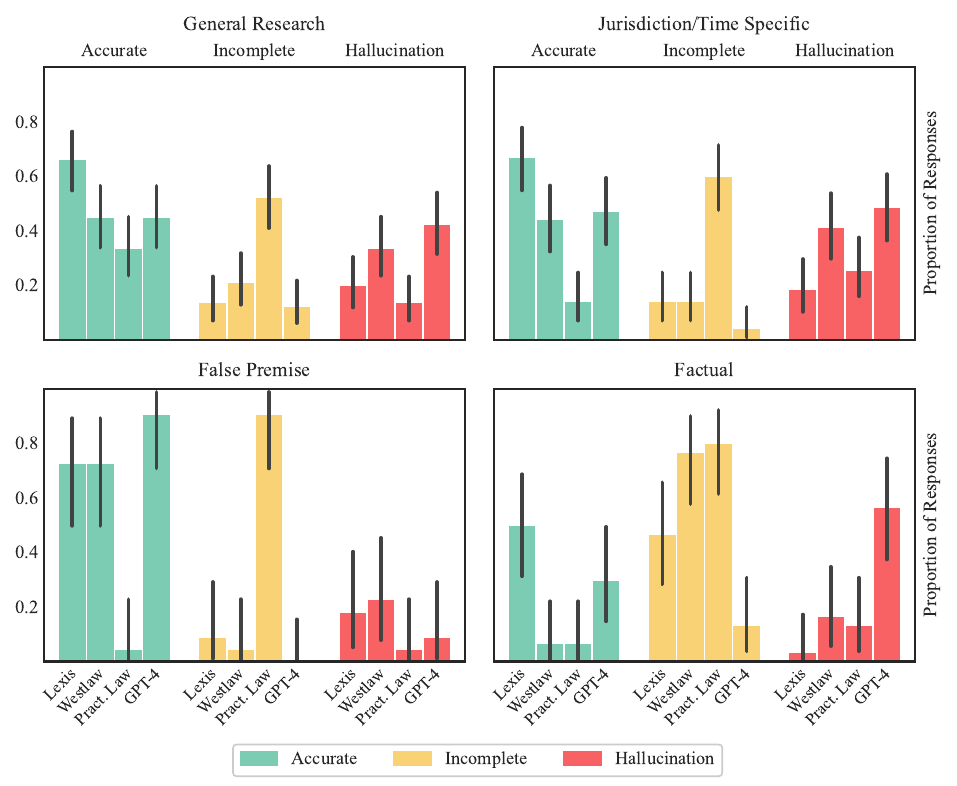}
        \caption{Response evaluations broken down by question category. We show the accuracy (green), incompleteness (yellow), and hallucination (red) rate for each question category. Vertical bars denote 95\% confidence intervals. This figure shows that hallucinations are not driven by an isolated category and persist across task types and questions, such as bar exam and appellate litigation issues.}
        \label{fig:categories_facet}
    \end{figure*}

Commercially-available RAG-based legal research tools still hallucinate. Over 1 in 6 of our queries caused Lexis+ AI and Ask Practical Law AI to respond with misleading or false information. And Westlaw hallucinated substantially more---\emph{one-third of its responses} contained a hallucination.
    
On the positive side, these systems are less prone to hallucination than GPT-4, but users of these products must remain cautious about relying on their outputs. 

The left panel of Figure~\ref{fig:stacked_responsive} provides a breakdown of response types across the four products. Lexis+ AI's answers are accurate (i.e., correct and grounded) for 65\% of queries, compared to much lower accuracy rates of 41\% and 19\% by by Westlaw and Practical Law AI, respectively.  The right panel of Figure~\ref{fig:stacked_responsive} also provides the hallucination rate when an answer is responsive, showing that Lexis+ AI appears to have a statistically significantly lower hallucination rate than Westlaw and Thomson Reuters, even conditional on a response.

Figure~\ref{fig:categories_facet} also breaks down these statistics by query type. We observe that, while hallucination rates are slightly higher for jurisdiction and time specific questions, they remain high for general legal research questions, such as questions posed on the bar exam. Accuracy rates are highest on ``false premise'' questions---in which the query contains a mistaken understanding of law---and lower on the categories which represent real-world use by attorneys.

Westlaw's high hallucination rate is driven by several kinds of errors (as discussed further in Section~\ref{sec:aiar}), but we note that it is also the system which tends to generate the longest answers. Excluding refusals to answer, Westlaw has an average word length of 350 (SD = 120), compared to 219 (SD = 114) by Lexis+ AI and 175 (SD = 67) by Ask Practical Law AI.\footnote{This is based on a simple word count separating based on space. }  With longer answers, Westlaw contains more falsifiable propositions and therefore has a greater chance of containing at least one hallucination. Lengthier answers also require substantially more time to check, verify, and validate, as every proposition and citation has to be independently evaluated. 

Responsiveness differs dramatically across systems. As shown in Figure \ref{fig:stacked_responsive}, Lexis+ AI, Westlaw AI-AR, and Ask Practical Law AI provide incomplete answers 18\%, 25\% and 62\% of the time, respectively.  The low responsiveness of Ask Practical Law AI can be can be explained by its more limited universe of documents. Rather than connecting its retrieval system to the general body of law (including cases, statutes, and regulations), Ask Practical Law AI draws solely from articles about
legal practice written by its in-house team of lawyers.

On the other hand, the Westlaw and Lexis retrieval systems are connected to a wider body of case law and primary sources. This means that they have access to all the documents that are in principle necessary to answer any of our questions. Both systems often offer high-quality responses. In one instance, Lexis+ AI pointed to a false premise in one of our questions. The question \texttt{scalr-19} asked whether the six year statute of limitation applied to retaliatory discharge actions under the False Claims Act. The question was drawn from \emph{Graham County Soil \& Water Conservation District v.\ U.S.}, 559 U.S.\ 280 (2010), where the Court held that there was ambiguity. Congress moved thereafter to amend the statute  to clarify the statute of limitations. Lexis+ AI explained the mistaken premise, and cited the relevant, updated code section. Similarly, when prompted about the need for specific, proven ``teaching, suggestion, or motivation'' (TSM) that would have led a person of ordinary skill in the art to combine the relevant prior art for a finding of obviousness, AI-AR correctly responded by discussing the Supreme Court's decision in \emph{KSR v.\ Teleflex}, 550 U.S.\ 398 (2007), which rejected a rigid notion of the Federal Circuit's TSM test.

\begin{table*}[tp]
    \renewcommand{\arraystretch}{1.3}
    \centering
    \begin{tabular}{p{0.55\textwidth} p{0.45\textwidth}}
        \toprule
        \textbf{Westlaw AI-AR Hallucination} & \textbf{Explanation} \\
        \midrule
        \noalign{\global\arrayrulewidth=0.01mm}
        \arrayrulecolor{gray}
        1. ``[W]hen the lender receives the collateral that secured the fraudulent loan, this is \textcolor{red}{considered a return of `any part' of the loan} money \dots  This was established in the Supreme Court case Robers v.\ U.S.'' & \textcolor{blue}{\emph{Robers} held precisely the opposite}: ``the phrase `any part of the property \dots returned' refers to the property the banks lost \dots and not to the collateral.'' 572 U.S. 639, 642 (2014).\\ 
        \hline
        2. ``D.M.\ v.\ State \dots has been overruled by Davis v.\ State. Also, the case \textcolor{red}{Millbrook v.\ U.S. was reversed} by the same case at a later date.'' & \emph{Millbrook v.\ United States} is a \textcolor{blue}{U.S. Supreme Court decision that controls on federal questions.} 569 U.S. 50 (2013). \textcolor{blue}{The Nebraska Supreme Court did not cite, much less `reverse,' it in \emph{Davis v.\ State}}. 297 Neb. 955 (2017).\\
        \hline
        3. ``[A] \textcolor{red}{paragraph from the Federal Rules of Bankruptcy Procedure [FRBP]}, Rule 4007 states that the deadlines set by bankruptcy rules governing the filing of dischargeability complaints are \textcolor{red}{jurisdictional}.'' & \textcolor{blue}{There is no such paragraph in the FRBP} and 4007's deadlines are unlikely to be jurisdictional under \emph{Kontrick v.\ Ryan}, 540 U.S. 443, 447--48, 448 n.3 (2004).\\
        \hline
        4. ``The Supreme Court of Alaska \dots noted \dots
        in the absence of physical injury, the losses for claimed emotional damage [from a misdiagnosis] \textcolor{red}{should not be permitted} for legal and public policy reasons (Chizmar v.\ Mackie, 896 P.2d 196 (1995)).''& The Supreme Court of Alaska in fact \textcolor{blue}{reversed this trial court holding} (that losses are not permitted on public policy grounds) and \textcolor{blue}{allowed} the question to proceed to the jury. 896 P.2d at 204-05.\\ 
        \hline
        5. ``[I]n the Doo v.\ Packwood case, a restrictive  covenant precluding the operation of a grocery store in a building conveyed was found to be potentially \textcolor{red}{inequitable and unenforceable} \dots (Doo v.\ Packwood, 265 Cal.App.2d 752 (1968)).'' & \textcolor{blue}{The \emph{Doo} court in fact held the opposite}. It reversed the trial court's finding that the covenant was unenforcable. 265 Cal.App.2d at 756, 759.\\
        \hline
        6. ``[T]he FCC's new policy regarding \textcolor{red}{`fleeting expletives' is arbitrary and capricious} under the Administrative Procedure Act.'' & \textcolor{blue}{The Supreme Court in fact found the opposite}: the fleeting expletives policy was \underline{not} arbitrary and capricious. FCC v.\ Fox, 556 U.S. 502 (2009). \\
        \hline
        7. ``In the Reid v.\ Fargo case, \textcolor{red}{there is no mention or reference made to the Irvine v.\ The Hesper case}.'' & \textcolor{blue}{The \emph{Reid} court expressly relied on the \emph{Hesper} case}. 241 U.S. 544, 548 (1916). \\
        \hline
        8. ``In the case of Bragg v.\ Linden Research, Inc., \textcolor{red}{the \underline{court} recognized participants' full intellectual property protection} for the digital content they created or owned in the game Second Life \dots Bragg v.\ Linden Research, Inc., 487 F.Supp.2d 593 (2007).'' & \textcolor{blue}{The court took no such position and was only describing actions by the \underline{defendant} (Linden)}: ``Linden announced that it would recognize participants' full intellectual property protection.'' 487 F.Supp.2d at 595.\\
        \hline
        9. ``In Washington Legal Foundation v.\ Legal Foundation of Washington, the \textcolor{red}{Court of Appeals for the Ninth Circuit held} that the interest generated by IOLTA pooled trust accounts is the property of the clients, and the appropriation of that interest for charitable purposes was a \textcolor{red}{per se `taking'} within the meaning of the Fifth Amendment.'' & This statement \textcolor{blue}{falsely credits the holding of a Ninth Circuit panel instead of the superseding holding of the Ninth Circuit en banc court}, which overruled the panel. 271 F.3d 835, 857-58 (applying ad hoc, not per se takings analysis).\\
        \hline
        10. ``\textcolor{red}{18 U.S.C.A. § 3583} \dots \textcolor{red}{was recognized as unconstitutional} by United States v.\ Winter Rose Old Rock. & The \emph{Winter Rose} court in fact \textcolor{blue}{rejected the defendant's claim that § 3583 was unconstitutional}, 76 F.4th 1314, 1316 (10th Cir. 2023), and only discussed and distinguished \emph{Haymond}, 588 U.S. 634 (2019) (4-1-4 decision with judgment that specific application of § 3583\textbf{(k)} was unconstitutional).  \\
        
         \arrayrulecolor{black}
        \bottomrule
    \end{tabular}
    \caption{Ten examples of hallucinations in Westlaw's AI-Assisted Research responses, with explanations for why they are coded as hallucinations.}\label{tab:aiar}
\end{table*}

 \subsection{Hallucinations Can Be Insidious}
    \label{sec:aiar}

These systems can be quite helpful when they work. But as we now illustrate in detail, their answers are often significantly flawed.
We find that these systems continue to struggle with elementary legal comprehension: describing the holding of a case \citep{zheng2021does}, distinguishing between legal actors (e.g., between the arguments of a litigant and the holding of the court), and respecting the hierarchy of legal authority. Identifying these misunderstandings often requires close analysis of cited sources. These vulnerabilities remain problematic for AI adoption in a profession that requires precision, clarity, and fidelity.

Tables~\ref{tab:aiar}, \ref{tab:lexis}, and \ref{tab:practical} provide examples of hallucinations in the Westlaw, Lexis, and Practical Law systems, respectively.\footnote{The number of examples reported are roughly proportional to the relative hallucination rates between tools.} In each example, our detailed analysis of responses and cited cases reveals a serious inaccuracy and hallucination in the system response. The following sections refer to examples in these tables to illustrate different failure modes in legal RAG systems.

\paragraph{Misunderstanding Holdings.} Systems do not seem capable of consistently making out the holding of a case. This is a serious issue, as legal research relies centrally on distinguishing the holding from other parts of the case. Table~\ref{tab:aiar} rows 1, 4, 5, and 6 provide examples of when Westlaw states a summary that is the direct opposite of the actual holding of a case, including case by the U.S.\ Supreme Court. For instance, Westlaw states that collateral is considered a return of ``any part'' of the loan, indicating that this was established by the Supreme Court in \emph{Robers v.\ U.S.}, but \emph{Robers} held the exact opposite (Table~\ref{tab:aiar} row 1).  An in another response, Lexis+ AI recites Missouri legislation criminalizing unauthorized camping on state-owned lands. But that legislation comes from the statement of facts and analysis, and in the cited case, the Missouri Supreme Court actually held that legislation unconstitutional (Table~\ref{tab:lexis}, row 3).

\paragraph{Distinguishing Between Legal Actors.} Systems can fail to distinguish between arguments made by litigants and statements by the court. In one example, Westlaw attributes an action of the defendant to the court (Table~\ref{tab:aiar} row 8) and in another it stated that a provision of the U.S.\ Code was found unconstitutional by the 10th Circuit, when in fact the 10th Circuit rejected that argument by the defendant (Table~\ref{tab:aiar}, row 10). 

\paragraph{Respecting the Order of Authority.} All models strain in grasping hierarchies of legal authority.  This is crucial, as courts often discuss similar propositions that may be in tension. When sources conflict, a complex system of precedence and hierarchy determines governing law. Sorting through different sources to find the authoritative ones requires legal ``background knowledge'' about the way that different courts interact in different jurisdictions, and even systems with direct access to case law can fail to adhere to these legal hierarchies. For example:

\begin{itemize}
\item Westlaw asserts that a U.S.\ Supreme Court case was reversed by the \emph{Nebraska Supreme Court} on a matter of federal law. That is not possible in the U.S.\ legal system, and in fact the Nebraska Supreme Court did not so much as cite the Supreme Court case in question (Table~\ref{tab:aiar} row 2).
\item Westlaw confuses holdings between different levels of courts (Table~\ref{tab:aiar} rows 5, 9). In row 9, for instance, Westlaw properly states the holding of the Ninth Circuit panel, but improperly attributes it to the Ninth Circuit sitting en banc, which actually overruled the panel on that issue. 
\item Lexis+ AI fails to distinguish between district and appellate courts. In Table~\ref{tab:lexis} row 1, Lexis+ AI transmogrifies a district court recitation of the trial court standard for awarding attorney's fees into a patently incorrect standard of \emph{appellate review} of attorney's fees (incorrectly stating that an appeals court may disturb attorney's fees ``as long as they provide reasoning''). 
\item In Table~\ref{tab:lexis} row 2, Lexis+ AI describes a rule established in \emph{Arturo D.} as good law, with citation to the case that actually overrules \emph{Arturo D.} 
\end{itemize}

We note one additional area where systems struggle with orders of authority. In numerous instances, we observed the Westlaw system stating a proposition based on an overruled or reversed case, \textit{without} citing the case. These errors may stem from design choices: Westlaw may be adding citations in a second pass, after generating the statement, while suppressing the citation of cases that receive a ``red flag'' under its KeyCite system.\footnote{Per Westlaw, a red flag indicates that a case ``is no longer good law for at least one of the points of law it contains''\citep{westlaw:keycite}. In our labeled sample, we were not able to observe such cases being cited, though they were sometimes discussed without citation.} For instance, when prompted about the equity clean-up doctrine, which allows courts of equity to decide legal and equity issues when it has jurisdiction over the equity issues, AI-AR properly cites the rule, but then notes, ``However, this general rule does not apply when the facts relied on to sustain the equity jurisdiction fail of establishment.'' This statement is unaccompanied by an in-text citation; the language appears only in a search result below the response, in a Missouri case\footnote{State ex rel. Leonardi v. Sherry, No. ED 82789, 2003 WL 21384384, at *1 (Mo. Ct. App. June 17, 2003).} that was overruled on that issue by the Missouri Supreme Court.\footnote{See State ex rel. Leonardi v. Sherry, 137 S.W.3d 462, 472 (Mo. 2004) (``The dissenting opinion apparently would cling to the inefficient and wasteful need for a second trial at law if equity `fails of establishment' in the initial request for equitable relief.'').} We believe this suppression behavior can be dangerous---it impedes verification of the claims most likely to be false. %

\paragraph{Fabrications.} The systems we test occasionally generate text that is unrelated or deviates materially from retrieved documents.

\begin{itemize}
\item Westlaw generates provisions of law that do not exist. For instance, it asserts that the Federal Rules of Bankruptcy (FRBP) state that deadlines are jurisdictional, which is not a statement contained in the FRBP text at all (Table~\ref{tab:aiar} row 3). (The hallucination seems to emanate from a retrieved 1996 bankruptcy court case, which is also likely invalid under the Supreme Court's \emph{Kontrick} decision, which found that bankruptcy deadlines are not jurisdictional.) 
\item Westlaw misinterprets the Supreme Court's specific holding on a statutory \emph{subsection} as the 10th Circuit finding the entire statutory section unconstitutional, when in fact the 10th Circuit rejected the defendant's claim of unconstitutionality (Table~\ref{tab:aiar} row 10). 
\item Lexis+ AI attributes a description of the equity clean-up doctrine to a case that only discusses the ``clean hands'' doctrine (Table~\ref{tab:lexis} row 4).
\end{itemize}

\begin{table*}[tp]
    \renewcommand{\arraystretch}{1.3}
    \centering
    \begin{tabular}{p{0.55\textwidth} p{0.45\textwidth}}
        \toprule
        \textbf{Lexis+ AI Hallucination} & \textbf{Explanation} \\
        \midrule
        \noalign{\global\arrayrulewidth=0.01mm}
        \arrayrulecolor{gray}

        1. ``In Vadatech v.\ Overnight [2016 Nev.\ Dist.\ LEXIS 1619], the court awarded attorney's fees and costs to the prevailing party. It cited Nevada Supreme Court precedent that \textcolor{red}{appellate courts can modify fee awards as long as they provide reasoning}, without needing to find an abuse of discretion.'' & \textcolor{blue}{Vadatech, a district court opinion, said no such thing} and instead recited the \textcolor{blue}{trial court} standard for granting attorney's fees---set by the Nevada Supreme Court---not the standard of \textcolor{blue}{appellate} modification of such fees. 
         \\
         \hline

        2. ``If a driver is unable to produce identification documents or produces documents that appear to be false, a limited search of places in the vehicle where the driver may have hidden a wallet is reasonable. In the \textcolor{red}{case of Arturo D., the court upheld the warrantless vehicle search} when the driver failed to produce the required documentation People v.\ Lopez, 8 Cal.\ 5th 353.'' & \textcolor{blue}{\emph{Lopez} actually overturned \emph{Arturo D.}}. Lexis+ AI describes \emph{Arturo D.} as if it is good law and supported by \emph{Lopez}. \\
        \hline

        3. ``The new section \textcolor{red}{67.2300 of TAFP HB 1606}, which was signed into law by Governor Parson \dots \textcolor{red}{makes the act of unauthorized sleeping and camping on state-owned lands a class C misdemeanor}. Byrd v. State, 679 S.W.3d 492.'' & HB 1606 was in fact \textcolor{blue}{struck down by the Missouri Supreme Court} in the cited case \textit{Byrd v.\ State}, 679 S.W.3d 492 (Mo.\ 2023). This detail is not mentioned in Lexis+ AI's response.\\
        \hline
        4. ``The \textcolor{red}{equity clean-up doctrine is a self-imposed ordinance that closes the doors of a court of equity to one tainted with an inequitableness} or bad faith relative to the matter in which he seeks relief. \dots  EEOC v. Recruit U.S.A., Inc., 939 F.2d 746'' &  This statement \textcolor{blue}{confuses the equity clean-up doctrine with the ``clean hands'' doctrine}. \emph{EEOC v.\ Recruit U.S.A.}, 939 F.2d 746, 752 (9th Cir.\ 1991).\\
        \hline
        5. ``The Telephone Consumer Protection Act of 1991 (TCPA) grants federal courts exclusive jurisdiction over actions brought by the state attorneys general, but \textcolor{red}{places exclusive jurisdiction over private civil actions in state courts}.'' & The Supreme Court in fact found \textcolor{blue}{concurrent state and federal court jurisdiction} over private suits under the TCPA. \emph{Mims v.\ Arrow Financial Services}, 565 U.S.\ 368 (2012).\\
        \hline
      6. ``The Eleventh Circuit \textcolor{red}{has not definitively ruled on the causation standard for Family and Medical Leave Act} (FMLA) retaliation claims. Some courts have applied the "but-for" causation standard, which requires a showing that the adverse action would not have occurred but for the protected activity  Lapham v. Walgreen Co., 88 F.4th 879 \dots However, other courts have applied a motivating factor or negative factor standard\dots{} Lee v. Yellowpages.Com LLC, 2019 U.S. Dist. LEXIS 250502, Nelson v. Americold Logistics, LLC, 2020 U.S. Dist. LEXIS 63611.'' & Lexis claims the standard is not definitive, but \textcolor{blue}{\emph{Lapham} is the most recent controlling case}, and the other cases contesting the issue are older inferior court rulings.
      \\
         \arrayrulecolor{black}
        \bottomrule
    \end{tabular}
    \caption{Six examples of hallucinations in Lexis+ AI responses with explanations for why they are coded as hallucinations.}\label{tab:lexis}
\end{table*}

\begin{table*}[tp]
    \renewcommand{\arraystretch}{1.3}
    \centering
    \begin{tabular}{p{0.55\textwidth} p{0.45\textwidth}}
        \toprule
        \textbf{Practical Law Hallucination} & \textbf{Explanation} \\
        \midrule
        \noalign{\global\arrayrulewidth=0.01mm}
        \arrayrulecolor{gray}

        1. ``Based on 28 U.S.C § 2680(h), the United States does waive its sovereign immunity for certain intentional torts committed by federal investigative or law enforcement officers acting within the scope of their employment. However, \textcolor{red}{this waiver applies only if these officers are acting in their capacity to search, seize evidence, or make arrests} for federal crimes [Federal Tort Claims Act (FTCA): Exceptions and Defenses]'' & The cited Practical Law resource states: \textcolor{blue}{``However, the tortious act need not occur while the officer is searching, seizing, or arresting for the proviso to apply (Millbrook v. United States, 569 U.S.\ 50, 55-57 (2013)''} \\
        \hline

        2. ``Yes, a person subject to domestic-violence-related restraining orders \textcolor{red}{can be prohibited from purchasing a firearm in the Fifth Circuit}. [State and Local Regulation of Firearms Under the Second Amendment]'' & The Practical Law article actually states: ``Federal law prohibits a person who is subject to a domestic violence restraining order from possessing a firearm\dots However, \textcolor{blue}{the US Court of Appeals for the Fifth Circuit has held that this law violates the Second Amendment} (United States v.\ Rahimi, 61 F.4th 443 (5th Cir.\ 2023)''\\
        \hline

        3. ``A dying declaration is admissible in federal court under the Federal Rules of Evidence (FRE) 804(b)(2) \dots This provision is commonly invoked in homicide prosecutions, where the victim's last words about the identity or description of their killer are introduced into evidence. However, \textcolor{red}{it could theoretically apply to any case where these specific conditions are met}.'' & The dying declaration exception is not just ``commonly'' invoked in homicide prosecutions; FRE 804(b)(2) specifically \textcolor{blue}{limits the admissibility of dying declarations to cases involving ``a prosecution for homicide or in a civil case,''} not any case.\\

        \hline

        4.``The \textcolor{red}{undue burden standard applies in abortion cases} when assessing the constitutionality of state regulations on abortion before a fetus becomes viable. \dots [
Health Plan Coverage of Surgical and Medication Abortion and Related Services,
 Group Health Plan Coverage Considerations for Employers After the Overturning of Roe v. Wade, Employee Assistance Program (EAP) Compliance]'' & \textcolor{blue}{Dobbs v.\ Jackson Women's Health, 597 U.S.\ 215 (2022) overturns \emph{Casey} and the undue burden standard.} All three Practical Law articles discuss \emph{Roe} and \emph{Casey}'s reversal, with one title including the phrase ``Overturning of \emph{Roe}.'' \\
        
         \arrayrulecolor{black}
        \bottomrule
    \end{tabular}
    \caption{Four examples of hallucinations in Thomson Reuters's Ask Practical Law AI response, with explanations of why they are coded as hallucinations. The Practical Law documents cited are named in square brackets.}
    \label{tab:practical}
\end{table*}

\subsection{A Typology of Legal RAG Errors}
\label{sec:failure}

Interpreting why an LLM hallucinates is an open problem \citep{Ji2023, zou2023representation}. While it is possible to identify
correlates of hallucination \citep{Dahl2024}, it is hard to conclusively explain why a model
hallucinates on one question but not another, or why one model hallucinates where another does not.

RAG systems, however, are composed of multiple discrete components \citep{Gao2024}. While each piece may be a black box, due to the lack of documentation by providers, we can partially observe the way that information moves between them. Lexis+ AI, Ask Practical Law AI, and AI-AR each show the list of documents which were retrieved and given to the model (though not exactly which pieces of text are passed in). Consequently, comparing the retrieved documents and the written response allows us to develop likely explanations for the reasons for hallucination.

In this section, we present a typology of different causes of RAG-related hallucination that we observe in our dataset. Other analyses of RAG failure points identify a larger number of distinct failure points~\citep{Barnett2024, Chen2024}. Our typology collapses some of these, since we focus on broader causes that can be identified using the limited information we have about the systems we test. Our typology also introduces new failure points unique to the legal context that have not previously been considered in analyses of general-purpose RAG systems. Evaluations of general purpose RAG systems often assume that all retrievable documents (1) contain true information and (2) are authoritative and applicable, an assumption that is not true in the legal setting \citep{Barnett2024, Chen2024}.\footnote{\cite{Chen2024} consider the possibility of retrievable documents that contain false information. However, its evaluation focuses on a significantly simplified setting that is not applicable to the complexity of legal use cases.} Legal documents often contain outdated information, and their relevance varies by jurisdiction, time period, statute, and procedural posture. Determining whether a document is binding or persuasive often  requires non-trivial reasoning about its content, metadata, and relationship with the user query.

This typology is intended to be useful to both legal researchers and AI developers. For legal researchers, it illustrates some pathways to incorrect outputs, and highlights specific areas of caution. For developers, it highlights areas for improvement in these tools. The categories that we present are not mutually exclusive; the failures we observe are often driven by multiple causes or have unclear causes. Table \ref{tab:hallucination_cause} compares the prevalence of different hallucination causes in our typology. Because these are closed systems, we are not able to clearly identify a single point of failure for each hallucination.

\begin{table*}[h]
    \begin{center}
        \begin{tabular}{lrrr}
\toprule
\textbf{Contributing Cause} & \textbf{Lexis} & \textbf{Westlaw} & \textbf{Pract. Law} \\
\midrule
Naive Retrieval & 0.47 & 0.20 & 0.34 \\
Inapplicable Authority & 0.38 & 0.23 & 0.34 \\
Reasoning Error & 0.28 & 0.61 & 0.49 \\
Sycophancy & 0.06 & 0.00 & 0.03 \\
\bottomrule
\end{tabular}

    \end{center}
    \caption{This table shows prevalence of different contributing causes among all hallucinated responses for each model. Because the types are not mutually exclusive, the proportions do not sum to 1.}
    \label{tab:hallucination_cause}
\end{table*}
\textbf{Naive retrieval.} Many failures in the three systems stem from poor retrieval---failing to find the most relevant sources available to address the user's query. For instance, when asked to define the ``moral wrong doctrine,'' a doctrine pertaining to mistake-of-fact instructions in criminal prosecutions for morally wrongful acts (\texttt{doctrine-test-177}), Lexis+ AI relies on a source which defines moral \emph{turpitude}, a legal term of art with a seemingly similar but actually unrelated meaning.

Part of the challenge is that retrieval itself often requires legal reasoning.
As Section \ref{sec:rag_limitations} discusses, legal sources are not composed of unambiguous facts.
Lawyers are often taught to analyze situations with an IRAC framework---first identify the issue (I) and governing legal rule (R), then analyze (A) the facts with that rule to arrive at a conclusion (C) \citep{Guha2023}. For example, \texttt{bar-exam-96} asks whether an airline's motion to dismiss should be granted in a wrongful death suit arising out of a plane crash. Ask Practical Law AI retrieves sources discussing motions to dismiss in various contexts such as bankruptcy and patent litigation. But correctly answering this question requires identifying the true underlying issue as being one about \emph{tort negligence}, not general procedures for motions to dismiss. Thomson Reuters's tool likely errs because it fails to perform this analytical step prior to querying its database, thereby ending up with sources pertaining to the wrong issue.

\textbf{Inapplicable authority.} An inapplicable authority error occurs when a model cites or discusses a document that is not legally applicable to the query. This can be because the authority is for the wrong jurisdiction, wrong statute, wrong court, or has been overruled. This kind of error is uniquely important and prevalent in the legal setting, and has not been explored as thoroughly in prior literature \citep{Barnett2024, Gao2024}. One example is Lexis+ AI's response to \texttt{scalr-15}. This question asks about certain deadlines under Bankruptcy Rule 4004, but the model describes and cites a case about tax court deadlines under 26 U.S.C.S. § 6213(a) instead. This could be because the excerpt of the case that is given to the model does not include key information, or because the model was given that information and ignored it. Because it is not possible to see exactly what information is available to the model, it is not possible to say precisely where the error occurs.

\textbf{Sycophancy.} LLM assistants have been found to display ``sycophancy,'' a tendency to agree with the user even when the user is mistaken \citep{Sharma2023}. While sycophancy can cause hallucinations \citep{Dahl2024}, we found that Lexis+ AI, AI-AR, and GPT-4 were quite capable at navigating our false premise queries, and often corrected the false premise without hallucination. For example, \texttt{false-holding-statements-108} asks for a case showing that due process rights can be violated by negligent government action. Lexis+ AI steers the user towards the correct answer, stating that intentional interference can violate due process, and that negligent interference cannot, supporting these propositions with case law. Ask Practical Law AI also seldom hallucinated in this category, but refused to answer at all in the overwhelming majority of queries.

\textbf{Reasoning errors.} In addition to the more complex behaviors described above, LLM-based systems also tend to make elementary errors of reasoning and fact. The legal research systems we test are no exception. We observe such errors most frequently in Westlaw; though retrieved results often seemed relevant and helpful, the model would not always correctly reason through the text to arrive at the correct conclusion. In one instance (Table~\ref{tab:aiar} row 8), AI-AR describes a district court decision as ``recogniz[ing] participant's full intellectual property protection for the digital content they created or owned in the game Second Life\dots'' But as the passage cited by the model makes clear, the court held no such thing. It was describing the statements of the \emph{defendant}, and the language model made a simple factual error in describing the passage given to it.

\section{Limitations}
\label{sec:limits}

While our study provides critical information about widely deployed AI tools in legal practice, it comes with certain limitations. 

First, our evaluation is limited to three specific products by LexisNexis, Thomson Reuters, and Westlaw. The legal AI product space is growing rapidly with many startups (e.g., Harvey, Vincent AI) \citep{Ma2024}. Access to these emerging systems is even more restricted than to the services offered in LexisNexis and Westlaw, making evaluation exceptionally challenging.\footnote{Even AI-Assisted Research was exclusively available to law firms when we initially conducted the evaluation of Lexis+ AI and Ask Practical Law AI \citep{westlaw:avoids}.} That said, our approach provides a common benchmark that can be deployed for similar systems as they become available. 

Second, our evaluation only captures a point in time. 
Even over the course of our study, we noticed the responses of these systems---particularly Lexis+ AI---evolve over time. While these changes may improve responses, we note that benchmarking, evaluation, and supervision remain difficult when a model changes over time \citep{chen2023chatgpt}.\footnote{Indeed, even presenting the same query to these models may yield different answers each time, as the text decoding process may not be set to be deterministic (e.g., via the temperature parameter). GPT-4, for instance, is known not to be deterministic. It is also not clear what retrieval parameters (e.g., similarity threshold or top-$k$ value) are used, impeding consistent analysis of the model.} This is compounded by uncertainty over whether such differences are driven by changes in the base model (e.g., GPT-4) or by engineering by the legal technology provider. More generally, a fundamental concern for the evaluation of LLMs lies in test leakage---because language models are trained on all available data, they may memorize data that is used for evaluation \citep{li2024task, oren2023proving, deng2023investigating}. That is a particularly challenging concern when the only mechanism for accessing legal AI tools is by sending test prompts to providers themselves. Even if providers fix the discrete errors noted above, that may not mean that the problems we identify have been solved in general.\footnote{For instance, OpenAI appeared to patch its system to prevent adversarial attacks with specific suffixes discovered in \citet{zou2023universal}, but the underlying vulnerability may still persist. As one of the authors of that study \href{https://x.com/andyzou_jiaming/status/1684766181381812225}{noted}, ``Companies like OpenAI have just patched the suffixes in the paper, but numerous other prompts acquired during training remain effective. Moreover, if the model weights are updated, repeating the same procedure on the new model would likely still work.''} 

Third, while we have been able to design an effective evaluation framework for chat-based interfaces, the evaluation for more specified generative tasks is still evolving. LegalBench \citep{Guha2023}, for instance, still requires manual evaluation of certain generative outputs, and we do not here assess Casetext CoCounsel's effectiveness at drafting open-ended legal memoranda. Developing benchmarks for the full range of legal tasks---e.g., deposition summaries, legal memoranda, contract review---remains an important open challenge for the field \citep{Kapoor2024}. 

Fourth, although we designed the first benchmark dataset, the sample size of 202 queries remains small in comparison to other evaluations such as \citet{Dahl2024}. There are two reasons for this. In contrast to general-purpose LLMs, which have open models or API access, LexisNexis, Thomson Reuters, and Westlaw restrict access to their interfaces.\footnote{See, for example, \S~2.2 of the LexisNexis Terms of Service \citep{LexisTos}, which prohibits programmatic access.} In addition, extensive \emph{manual} work is required to evaluate the results of each query, making it harder to scale automated evaluations. The trend toward LLM-based evaluations may address the latter obstacle, but the fact remains that the legal AI product space remains quite closed.  

Fifth, while we managed to develop a measurement protocol that yielded substantial agreement between human raters, we acknowledge that groundedness may exist on a spectrum. A citation, for instance, might point to a case that has been overruled, but that case might still be helpful to an attorney in starting the research process. In our setting, we coded such instances as misgrounded, but whether the model is helpful will still fundamentally have to be determined by use cases and evaluations that involve human interactions with the system. The range of failure points documented in Section~\ref{sec:failure} provides a more granular sense of the limitations of current AI systems. 

Sixth, some might argue that our benchmark dataset does not represent the natural distribution of queries. We designed our benchmark to reflect a wide range of query types and to constitute a challenging real-world dataset. Questions are ones that arise on the bar exam, that arise in appellate litigation, that present circuit splits, that present issues that are dynamically changing, and that were contributed by the legal community \citep{Guha2023}. The benchmark was designed to be challenging precisely because (a) those are the settings where legal research is needed the most, and (b) it responds to the marketing claims by providers. 
It is true that these may not represent all tasks for which lawyers turn to generative AI. Our estimate of the hallucination rate is not meant to be an unbiased estimate of the (unknown) population-level rate of hallucinations in legal AI queries, but rather to assess whether hallucinations have in fact been solved by RAG, as claimed. We show that hallucinations persist across the wide range of task types (see Figure~\ref{fig:overall}) and the full natural distribution of such queries is (a) only known to legal technology providers, (b) highly in flux given uncertainty about the appropriate use of AI in law, and (c) itself endogenous to assessments of reliability and marketing claims. 

Last, our primary goal is limited to assessing the hallucination rate, accuracy, and groundedness on emerging legal technology. These are central concepts to the trustworthiness of AI tools, but they are not the sole criteria for the quality and value of a legal research system. For instance, notwithstanding the many hidden hallucinations, the overall output of Lexis+ AI and AI-AR may still be quite valuable for distinct use cases (e.g., starting on a research thread). But evaluations like the one we designed here are critical to understanding these appropriate use cases. 

\section{Implications}
\label{sec:implications}

Excitement over the potential for AI to transform the practice of law is at an all-time high. On the demand side, lawyers fear missing out on the real gains in efficiency and thoroughness that new AI tools can offer. On the supply side, the companies developing these tools continue to market them as more and more powerful \citep{Markelius2024}. We agree that these tools are hugely promising \citep{Chien2024a, Choi2024}, but our research has important implications for both the lawyers using these products and the myriad of companies now marketing them.

\subsection{Implications for Legal Practice}

In the United States, all lawyers are required to abide by certain professional and ethical rules. Most jurisdictions have adopted a version of the Model Rules of Professional Conduct, which are issued by the American Bar Association \citep{AmericanBarAssociation2018}. Two of these rules bear directly on the integration of AI into law: Rule 1.1's duty of competence and Rule 5.3's duty of supervision \citep{Cyphert2021, Walters2019, Yamane2020}. Competence requires ``legal knowledge, skill, thoroughness and preparation'' (Rule 1.1); supervision requires ``reasonable efforts to ensure that the [non-lawyer's] conduct is compatible with the professional obligations of the lawyer'' (Rule 5.3).

In addition to these rules, the bar associations of New York \citeyearpar{ny-bar2024}, California \citeyearpar{ca-bar2023}, and Florida \citeyearpar{florida-bar2024} have all recently published more detailed guidance on how lawyers' ethical responsibilities intersect with their use of AI. For example, the New York State Bar Association's AI Task Force states that lawyers ``have a duty to understand the benefits, risks and ethical implications'' associated with the tools that they use \citeyearpar[57]{ny-bar2024}; similarly, the State Bar of California's Standing Committee on Professional Responsibility and Conduct implores lawyers to ``understand the risks and benefits of the technology used in connection with providing legal services'' \citeyearpar[1]{ca-bar2023}.

In other words, lawyers' ability to comply with their professional duties in both of these jurisdictions is contingent on access to \textit{specific} information about empirical risks and benefits of legal AI. Yet, so far, no legal AI company has provided this information. The New York State Bar Association points its members to a list of publications and fora that discuss matters related to AI in general \citeyearpar[76-77]{ny-bar2024}, but general knowledge is not the same as understanding the trade-offs of specific tools.

Indeed, our work shows that the risks and benefits associated with AI-driven legal research tools are different from those associated with general-purpose chatbots like GPT-4. As we discuss in Section~\ref{sec:results}, the tools we study in this article differ in responsiveness and accuracy, and these differences may even change over time within the same tool. The closed nature of these tools, however, makes it difficult for lawyers to assess when it is safe to trust them. Official documentation does not clearly illustrate what they can do for lawyers and in which areas lawyers should exercise caution. Thus, given the high rate of hallucinations that we uncover in this article, lawyers are faced with a difficult choice: either verify by hand each and every proposition and citation produced by these tools (thereby undercutting the efficiency gains that AI is promised to provide), or risk using these tools without full information about their specific risks and benefits (thereby neglecting their core duties of competency and supervision).

\subsection{Implications for Legal AI Companies}

Legal AI developers face dilemmas as well. On the one hand, these companies are subject to economic pressures to compete in an increasingly crowded market \citep{Ma2024}, pressures made more acute by the recent entry of previously copyrighted and proprietary data into the public domain \citep{Henderson2022, Ostling2024, TheLibraryInnovationLab2024}. On the other hand, like all businesses, they are also constrained by laws and regulations limiting the products they can lawfully offer and advertise. We flag two of these potential restrictions here.

First, companies must be careful not to overclaim or misrepresent the abilities of their AI products. As we discuss in Section~\ref{sec:introduction}, a number of legal AI providers are currently making claims about their products' ability to ``eliminat[e]''~\citep{casetext:eliminates} or  ``avoid'' hallucinations~\citep{westlaw:avoids}, yet, as we note in Section~\ref{sec:hallucination_definition}, these same companies are inconsistently using the term ``hallucination'' in ways that may not conform to users' expectations. Without additional precision about the exact mistakes that their tools purportedly avoid, companies may find themselves exposed to civil liability for unfair competition or false, misleading, or unsubstantiated claims. For instance, under Section 43(a) of the Lanham Act, 15 U.S.C. \S~1125, both customers and competitors alike may seek to recover for damages caused by such practices. The Securities and Exchange Commission has charged investment advisers with false and misleading claims about AI \citep{sec2024}, expressing concerns about ``AI washing'' by public companies \citep{grewal2024}, and the Federal Trade Commission, too, has warned about deceptive AI claims lacking scientific support \citep{ftc2023}. 

Second, legal AI providers must also be cautious about emerging theories of tort liability for AI-inflicted harms. This territory is less well-charted, but a developing scholarly literature suggests that developers who negligently release AI products with known defects may also face legal exposure \citep{vanderMerwe2024, Wills2024}. For example, one airline company in Canada has already been held liable for negligent misrepresentation based on output produced by its AI chatbot \citep{Rivers2024}. From theories of vicarious liability \citep{Diamantis2023}, to products liability \citep{Brown2023}, to defamation \citep{Volokh2023, Salib2024}, legal AI providers must carefully weigh the potential tort risks of releasing products with known hallucination problems.

\section{Conclusion}
\label{sec:conclude}

AI tools for legal research have not eliminated hallucinations. Users of these tools must continue to verify that key propositions are accurately supported by citations.

The most important implication of our results is the need for rigorous, transparent benchmarking and public evaluations of AI tools in law. In other AI domains, benchmarks such as the Massive Multitask Language Understanding \citep{Hendrycks2020} and BIG Bench Hard~\citep{srivastava2023beyond,suzgun-etal-2023-challenging} have been central to developing a common understanding of progress and limitations in the field. But in contrast to even GPT-4---not to mention open-source systems like Llama and Mistral---legal AI tools provide no systematic access, publish few details about models, and report no benchmarking results at all. This stands in marked contrast to the general AI field \citep{Liang2023}, and makes responsible integration, supervision, and oversight acutely difficult.  

We note that some well-resourced firms have conducted internal evaluations of products. Paul Weiss, a firm with over \$2B in annual revenue, for instance, has conducted an internal evaluation of Harvey, albeit with no published results or quantitative benchmarks \citep{Gottlieb2024}. This itself has distributive implications on AI and the legal profession, as  ``businesses are looking to well-resourced firms \dots to get some 
understanding of how to use and evaluate the new software'' \citep{Gottlieb2024}. If only well-heeled actors can even evaluate the risks of AI systems, claims of functionality \citep{RajiFallacy} and that AI can improve access to justice may be quite overstated \citep{Bommasani2022, Chien2024, Perlman2023, Tan2023}. 

That said, even in their current form, these products can offer considerable value to legal researchers compared to traditional keyword search methods or general-purpose AI systems, particularly when used as the first step of legal research rather than the last word. Semantic, meaning-based retrieval of legal documents may be of substantial value independent of how these systems then use those documents to generate statements about the law. The reduction we find in the hallucination rate of legal RAG systems compared to general purpose LLMs is also promising, as is their ability to question faulty premises. 

But until vendors provide hard evidence of reliability, claims of hallucination-free legal AI systems will remain, at best, ungrounded.

\section{Acknowledgments}
We thank Pablo Arredondo, Mike Dahn, Neel Guha, Sandy Handan-Nader, Peter Henderson, Pamela Karlan, Larry Moore, Julian Morimoto, Dilara Soylu, Andrea Vallebueno, and Lucia Zheng for helpful comments.

Authors have no conflicts to disclose. For transparency, CDM is an advisor to various LLM-related companies both individually and through being an investment advisor at AIX Ventures. 

\clearpage

\bibliographystyle{acl_natbib}
\bibliography{custom}

\clearpage

\appendix

\section{Complete Query Descriptions}\label{appendix:queries}

\subsection{General Legal Research}

    \subsubsection{Multistate Bar Exam}
    \begin{description}
    \item[Description] Questions from the multiple-choice multistate bar exam, reformatted as open-ended questions (i.e., no response choices given).
    \item[\# of Queries in Dataset] 20
    \item[Example] Arnold decided to destroy
    an old warehouse that he owned because the taxes on the
    structure exceeded the income that he could receive
    from it. He crept into the building in the middle of
    the night with a can of gasoline and a fuse and set the
    fuse timer for 30 minutes. He then left the building.
    The fuse failed to ignite, and the building was not
    harmed. Arson is defined in this jurisdiction as ``The
    intentional burning of any building or structure of
    another, without the consent of the owner.'' Arnold
    believed, however, that burning one's own building was
    arson, having been so advised by his lawyer. Has Arnold
    committed attempted arson?
    \item[Source] BARBRI practice bar exam questions \citep{barbri}.
    \item[Evaluation Reference] BARBRI answer key.
    \end{description}

    \subsubsection{Rule QA}
    \begin{description}
    \item[Description] Questions asking the model
    to describe a well-established legal rule. These rules
    sometimes represent the kind of legal ``background
    knowledge'' that does not always require a citation to a
    specific case. Other rules are tied to a specific civil or
    criminal statute. They are also the kind of question that a
    lawyer may ask when learning about a new area of the law,
    and the kind of question that is not easy to
    keyword-search.
    \item[\# of Queries in Dataset] 20
    \item[Example] What are the four fair use factors?
    \item[Source] Rule QA task in LegalBench \citep{Guha2023}.
    \item[Evaluation Reference] LegalBench answer key.
    \end{description}

    \subsubsection{Treatment (Doctrinal Agreement)}
    \begin{description}
    \item[Description] Questions about how one Supreme Court case treated another Supreme Court case that it cites.
    \item[\# of Queries in Dataset] 20
    \item[Example] How did Nassau Smelting \& Refining Works, Ltd. v.\ United States, 266 U.S. 101 (1924) treat United States v.\ Pfirsch, 256 U.S. 547 (1924)?
    \item[Source] Entries in a Shepard's Citations dataset for the     Supreme Court \citep{Fowler2007, Black2013}.
    \item[Evaluation Reference] Whether the model correctly characterizes the treatment of the cited case, e.g., as ``followed'', ``distinguished'', ``overruled,'' etc.
    \end{description}

    \subsubsection{Doctrine Test}
    \begin{description}
    \item[Description] Questions asking the model to define a well-known legal doctrine taught in standard black-letter courses like contracts, evidence, procedure, or statutory interpretation.
    \item[\# of Queries in Dataset] 10
    \item[Example] What is the near miss doctrine?
    \item[Source] Hand-curated.
    \item[Evaluation Reference] Our own domain knowledge.
    \end{description}
    
    \subsubsection{Question with Irrelevant Context}
    \begin{description}
    \item[Description] The Doctrine Test questions, but with some irrelevant context prepended, which is not related to the questions and which the model is expected to ignore.
    \item[\# of Queries in Dataset] 10
    \item[Example] Escheat is the passing of an interest in land to the state when a decedent has no will, no heirs, or devisees. In the United States, escheat rights are governed by the laws of each state. Probate is usually used to determine escheat rights. What is the near miss doctrine?
    \item[Source] We selected arbitrary definitions from Black's Law Dictionary and appended them to our doctrine test questions.
    \item[Evaluation Reference] Our own domain knowledge.
    \end{description}

\subsection{Jurisdiction or Time-specific}

    \subsubsection{SCALR}
    \begin{description} \item[Description] Questions presented
    in Supreme Court cases decided between 2000 and 2019. The
    questions are slightly rephrased to be suitable to ask an
    LLM. The task measures whether the AI system correctly
    identifies legal standards after recent changes in law
    (which typically take place when a Supreme Court case is
    decided). Unlike the LegalBench version of this task, which
    is multiple-choice for easier evaluation, this is presented
    as an open-ended task.
    \item[\# of Queries in Dataset] 30
    \item[Example] Did Congress divest the federal district courts of their federal-question jurisdiction under 28 U.S.C. § 1331 over private actions brought under the Telephone Consumer Protection Act?
    \item[Source] SCALR task in LegalBench (derived from the questions presented hosted on the Supreme Court’s website) \citep{Guha2023}.
    \item[Evaluation Reference] LegalBench answer key containing a holding statement describing the
    relevant SCOTUS case. Evaluators may also refer to Oyez, or
    check for any overruled cases if relevant.
    \end{description}
    
    \subsubsection{Circuit Splits}
    \begin{description}
    \item[Description] Questions testing whether the model correctly identifies the law in a specific circuit on a legal question that circuits disagree on.
    \item[\# of Queries in Dataset] 10
    \item[Example] To prove the ``haboring'' of undocument immigrants
    in the Sixth Circuit under 8 U.S.C. § 1324, must the
    prosecution prove that the defendant acted intentionally?
    \item[Source] Bloomberg Law articles on current circuit splits.
    \item[Evaluation Reference] Bloomberg Law articles' descriptions of the rule in the specified court, with checks for overruling cases if any are cited.
    \end{description}
    
    \subsubsection{Changes in Law}
    \begin{description}
    \item[Description] Questions about legal issues whose correct answer recently changed because of a recent circuit court ruling.
    \item[\# of Queries in Dataset] 10
    \item[Example] In the Eighth Circuit, can a private litigant sue under Section 2 of the Voting Rights Act?
    \item[Source] A variety of news articles on recent changes in law in circuit courts.
    \item[Evaluation Reference] Circuit opinions and news articles' descriptions of different sides in circuit splits.
    \end{description}
    
    \subsubsection{Local Rules}
    \begin{description}
    \item[Description] Questions about the Local Rules of Civil Procedure in various United States District Courts, limited to issues where the Local Rules diverge from the Federal Rules of Civil Procedure.
    \item[\# of Queries in Dataset] 20
    \item[Example] In the U.S. District Court for the Southern District of Indiana, how many days before serving a Rule 45 subpoena to a non-party must a party first serve notice of the proposed subpoena to the other parties?
    \item[Source] Hand-curated from websites of federal district courts.
    \item[Evaluation Reference] Local rules sourced from the websites of federal district courts.
    \end{description}

\subsection{False Premise}

    \subsubsection{False Holding Statements}
    \begin{description}
    \item[Description] Questions seeking a case in a particular jurisdiction standing for a proposition that is the precise opposite of what a case in that jurisdiction has held.
    \item[\# of Queries in Dataset] 10
    \item[Example] I'm looking for a case in Colorado which holds that pro se litigants can be awarded attorney's fees.
    \item[Source] We randomly selected legal documents from CourtListener. We excluded cases that did not have clearly written parenthetical holding statements \citep{parentheticals}; brief, judge-written summaries of the key holding of the case. We hand-constructed a question by negating the holding statement. We asked the model to identify a case in the relevant jurisdiction that held the opposite of what this real case held.
    \item[Evaluation Reference] Our query asks for support for a proposition that has been contradicted in a particular jurisdiction. Acceptable responses are that (1) no such case exists, (2) a case to the contrary exists (citing a case similar to the one that our question negates), or (3) such a case does exist, and our evaluation finds that the case in fact does supersede the opinion that our question was drawn from. We did not observe any instances of (3).
    \end{description}
    
    \subsubsection{Whimsical Holding Statements}
    \begin{description}
    \item[Description] Questions seeking a case standing for a proposition that is legally preposterous. (The model should respond that no such case exists.)
    \item[\# of Queries in Dataset] 10
    \item[Example] I'm looking for a case that stands for the proposition that a pedestrian can be charged with theft for absorbing sunlight that would otherwise fall on solar panels, thereby depriving the owner of the panels of potential energy.
    \item[Source] Hand-written.
    \item[Evaluation Reference] We ask to support a clearly false proposition. Acceptable responses are that (1) no such cases exists, (2) a case to the contrary exists, or (3) a description of the case supporting the closest reasonable proposition, while explaining that the original proposition is false.
    \end{description}

    \subsubsection{False Judicial Contributions}
    \begin{description}
    \item[Description] Questions asking about the legal career of Judge Luther A. Wilgarten, a fictional jurist \citep{Miner1989}.
    \item[\# of Queries in Dataset] 2
    \item[Example] What contributions has Judge Luther A. Wilgarten made to the legal field?
    \item[Source] Hand-curated, with thanks to Professor Pam Karlan for inspiration.
    \item[Evaluation Reference] Any described contribution is a hallucination; Judge Wilgarten is fictional.
    \end{description}

\subsection{Factual Recall}

    \subsubsection{Metadata: Citation}
    \begin{description}
    \item[Description] Questions seeking the well-formatted Bluebook citation of a given case.
    \item[\# of Queries in Dataset] 10
    \item[Example] Give me a Bluebook-formatted citation for Sears, Roebuck \& Co. v.\ Blade, decided by the California Court of Appeals in 1956.
    \item[Source] Randomly selected legal documents from CourtListener.
    \item[Evaluation Reference] Publicly available case data from CourtListener \citep{courtlistener}.
    \end{description}

    \subsubsection{Metadata: Year}
    \begin{description}
    \item[Description] Questions seeking the year that a given case was decided.
    \item[\# of Queries in Dataset] 10
    \item[Example] What year was Riley v.\ Albany Savings Bank, 103 N.Y. 669 (NY) decided?
    \item[Source] Randomly selected legal documents from CourtListener.
    \item[Evaluation Reference] Publicly available case data from CourtListener \citep{courtlistener}.
    \end{description}

    \subsubsection{Metadata: Author}
    \begin{description}
    \item[Description] Questions seeking the author of the majority opinion in a given case.
    \item[\# of Queries in Dataset] 10
    \item[Example] Who wrote the majority opinion in In Re Bebar, 315 F. Supp. 841 (E.D.N.Y 1970)?
    \item[Source] Randomly selected legal documents from CourtListener.
    \item[Evaluation Reference] Publicly available case data from CourtListener \citep{courtlistener}.
    \end{description}

\section{Running Queries}

We ran queries against Lexis+ AI and Thomson Reuters Practical Law AI by pasting the
complete text of each query into the chat box, without system message or other
text. We started a new conversation for each query, so no state was preserved.
We copied the complete text of each response and pasted it into our records. In-text citations were included in our copy, and we made an effort to copy the list of materials presented after the response, but these were not consistently captured.

\subsection{Queries Modified after Pre-registration}
\label{appendix:modified_queries}

During the pre-registration process, we noted that we retain the flexibility to make minor, non-substantive edits to our questions. Any changes that we made to our queries after pre-registration are enumerated here.

\begin{description}
    \item [\texttt{scalr-2}] We inserted the word ``specific'' in the question to more accurately describe the legal distinction drawn by the Supreme Court in the case.
    \item [\texttt{scalr-9}] We inserted the phrase ``reasonable probability'' in the question to more accurately describe the legal distinction drawn by the Supreme Court in the case.
    \item [\texttt{changes-in-law-74}] We replaced ``midwife'' with ``nurse practitioner'' to more accurately capture the effect of the relevant change in law.
    \item[\texttt{bar-exam-90}] The original query was formatted as a fill-in-the-blank (``the defendant's testimony is''), and we rephrased it to be a proper question (``is the defendant's testimony admissible?'').
    \item[\texttt{metadata-citation-130}] The original query was mistakenly truncated, and we corrected it to include the court and year, as all the other citation queries do.
    \item[\texttt{local-rules-191} to \texttt{local-rules-200}] The original questions said, for example, ``the Southern District of Indiana,'' which could be interpreted to refer to state courts in Indiana. The questions were about federal courts, so we edited all of these to say, e.g., ``the \emph{U.S. District Court for the} Southern District of Indiana.''
    
\end{description}

\renewcommand{\paragraph}[1]{\textit{#1}}

\section{Query Evaluation}

\label{appendix:labeling}

The below materials reproduce the annotation criteria we adhered to during evaluation of queries.

\subsection{Instructions}

Every response must be coded with two values: correctness and groundedness. Note that Partially Correct answers were collapsed into Correct answers during our final analysis, and Irrelevant / Unhelpful and Stock Refusal answers were collapsed into the category Refusal.

\begin{enumerate}
\def\labelenumi{\arabic{enumi}.}
\item Evaluate whether the response is correct, referring to the correctness rubric.
  \begin{enumerate}
  \def\labelenumii{\alph{enumii}.}
  \item Groundedness is only judged for correct or partially correct responses;
  \item If the correctness is Stock Refusal, Irrelevant/Unhelpful, or Incorrect then groundedness must be N/A
  \item If the question is a Bluebook citation question, then groundedness must be Grounded
  \item If none of the criteria above are met, proceed to step 2.
  \end{enumerate}
\item Evaluate groundedness, referring to the groundedness rubric.
\end{enumerate}

\subsection{Correctness Rubric}
For the below labels, we provide example responses to the hypothetical query: \emph{Do law enforcement officers in California have to inform drivers why they are being pulled over?}.

\subsubsection*{Correct}
The answer provides accurate information that is fully responsive to the query.

\emph{Example: Yes. AB 1542 goes into effect in 2024, which requires California police officers to inform drivers about the reason for the stop\ldots{}}

\subsubsection*{Partially Correct}
The answer contains no false propositions, but it does not address the substance of the question, or fails to include a piece of information relevant to the question.

\emph{Example: Yes, law enforcement officers in California are generally required to inform drivers why they are being pulled over. This requirement is part of the procedural norms that ensure transparency and fairness\ldots{}} (there is no mention of the relevant CA law)

\subsubsection*{Irrelevant/Unhelpful}
The response contains irrelevant or unhelpful information, not answering the question that is asked. However, it does not contain any false information or statements.

\emph{Example: The Fourth Amendment requires law enforcement officers to obtain a warrant prior to entering a suspect's home\ldots{}}

\subsubsection*{Stock Refusal}
The system provides a rote refusal to answer the question.

\emph{Example: The sources provided contain no information relevant to the query.}

\subsubsection*{Incorrect}
The response makes any false statement, whether material to the response or not.

\subsubsection*{Notes on Correctness}
\paragraph{Coding False Premise Questions}

For false premise questions, a response indicating that no relevant authority could be located is coded as Correct, and not Irrelevant/Unhelpful. However, a stock refusal without any such indication is coded as a Refusal.

\begin{itemize}
\item ``I cannot provide you with any information on this topic.'' (Refusal)
\item ``I cannot find any information on this topic.'' (Correct)
\item ``X case held the opposite to the premise presented.'' (Correct)
\end{itemize}

\paragraph{Coding Bluebook Citation Responses}
\begin{itemize}
\item We are strict Bluebookers. Accept only entirely compliant definitions; missing years, courts, or any information in the Bluebook standard citation is \textbf{incorrect}.
\item For example, if the parenthetical contains the year but not the court (where the court is required by \emph{The Bluebook}), that is incorrect.
\item A citation in which the year is off by one is incorrect
\end{itemize}

\subsection{Groundedness Rubric}
\subsubsection*{Grounded}
Every legal proposition which is material (i.e. relevant and non-trivial) to the query is supported by an applicable legal source. Indirect support is acceptable; i.e. a citation to a document which then cites an applicable document is grounded.

\subsubsection*{Ungrounded}
Every legal proposition which is material (i.e. relevant and non-trivial) to the query requires a citation to a source. If any material proposition is not supported by a citation, the response is ungrounded.

\subsubsection*{Misgrounded}
The system supports a proposition with a source which does not in reality support the proposition.

\subsubsection*{Fabricated}
The answer cites a source which does not exist.

\subsubsection*{Not Applicable}
Only coded when no factual propositions are present; only selected for Irrelevant/Unhelpful and Stock Refusal responses.

\subsubsection*{Notes on Groundedness}
\paragraph{Multiple Propositions, Single Source}
\begin{itemize}
\item A model may sometimes assert two distinct propositions and cite a single source at the end. If the single source supports both propositions, we consider that \textbf{grounded}. However, if both propositions are material to the user's query and only the latter proposition is supported by the source, the response is \textbf{ungrounded}.

  \begin{itemize}
  \item ``The Constitution protects the right to interracial marriage. It also protects the right to same-sex marriage. \emph{Obergefell v.\ Hodges}\ldots'' --- Grounded, because \emph{Obergefell} includes discussion of \emph{Loving v.\ Virginia} and its recognition of a right to interracial marriage
  \item ``The exclusionary rule prevents the admission of unlawfully obtained evidence. The Constitution protects the right to same-sex marriage. \emph{Obergefell v.\ Hodges} \ldots'' --- Ungrounded, because the source supports only the second proposition
  \end{itemize}
\end{itemize}

\begin{itemize}
\item A response can be both ungrounded and misgrounded, e.g. if Proposition 1 contains no support and Proposition 2 is incorrectly supported. In this case, the response is labeled with the most serious offense: Misgrounded.
\end{itemize}

\paragraph{Miscellaneous}
\begin{itemize}
\item If the primary (``correctness'') label of an example is irrelevant or unhelpful, then its secondary (``groundedness'') label should be N/A.
\item If the primary label of an example is incorrect, then the secondary label should be N/A.
\end{itemize}

\end{document}